\documentclass[conference]{IEEEtran}
\usepackage{amsmath,amsfonts}
\usepackage{algorithmic}
\usepackage{graphicx}
\usepackage{textcomp}
\usepackage{xcolor}
\usepackage{url}
\usepackage{adjustbox}
\usepackage{multirow}
\usepackage{enumitem}
\usepackage[algo2e,linesnumbered,ruled,vlined]{algorithm2e}
\usepackage{bm}
\usepackage{setspace}
\usepackage{cite}
\usepackage{tablefootnote}

\newcommand{\floor}[1]{\left\lfloor #1 \right\rfloor}
\newcommand{\ceil}[1]{\left\lceil #1 \right\rceil}

\newcommand{\argmin}{\operatornamewithlimits{argmin}}

\usepackage{amsthm}
\newtheorem{theorem}{Theorem}

\newlength\mylen

\usepackage[caption=false]{subfig}

\def\BibTeX{{\rm B\kern-.05em{\sc i\kern-.025em b}\kern-.08em
		T\kern-.1667em\lower.7ex\hbox{E}\kern-.125emX}}

\newcommand{\n}{SDT-GNN}
\newcommand{\algoname}{SPRING}

\begin{document}

\title{SDT-GNN: Streaming-based Distributed Training Framework for Graph Neural Networks}

\author{
\IEEEauthorblockN{Xin Huang\textsuperscript{$\ast$}, Weipeng Zhuo\textsuperscript{$\star$}, Minh Phu Vuong\textsuperscript{$\ast$}, Shiju Li\textsuperscript{$\dagger$}, \\ Jongryool Kim\textsuperscript{$\dagger$}, Bradley Rees\textsuperscript{$\ddagger$}, Chul-Ho Lee\textsuperscript{$\ast$}\vspace{1.5mm}}
\IEEEauthorblockA{\textsuperscript{*}Texas State University, 
\textsuperscript{$\star$}Beijing Normal–Hong Kong Baptist University, 
\textsuperscript{$\dagger$}SK hynix America, 
\textsuperscript{$\ddagger$}NVIDIA}
}

\maketitle

\begin{abstract}
Recently, distributed GNN training frameworks, such as DistDGL and PyG, have been developed to enable training GNN models on large graphs by leveraging multiple GPUs in a distributed manner. Despite these advances, their memory requirements are still excessively high, thereby hindering GNN training on large graphs using commodity workstations. In this paper, we propose \n{}, a streaming-based distributed GNN training framework. Unlike the existing frameworks that load the entire graph in memory, it takes a stream of edges as input for graph partitioning to reduce the memory requirement for partitioning. It also enables distributed GNN training even when the aggregated memory size of GPUs is smaller than the size of the graph and feature data. Furthermore, to improve the quality of partitioning, we propose \algoname{}, a novel streaming partitioning algorithm for distributed GNN training. We demonstrate the effectiveness and efficiency of \n{} on seven large public datasets. \n{} has up to $95\%$ less memory footprint than DistDGL and PyG without sacrificing the prediction accuracy. \algoname{} also outperforms state-of-the-art streaming partitioning algorithms significantly.
\end{abstract}

\section{Introduction}

Graph neural networks (GNNs)~\cite{ma2021deep, hamilton2020graph} have received significant attention due to their excellent performance for a wide range of graph-related learning tasks such as node classification, link prediction, and graph classification~\cite{defferrard2016convolutional,hamilton2017inductive,xu2018powerful,huang2025demystifying}. They have also been shown successful in other domains, such as natural language processing~\cite{yao2019graph,wu2023graph} and computer vision~\cite{qi2018learning,hu2018relation}. To facilitate their implementation and accelerate the model training process, there have been several GNN frameworks and libraries~\cite{FeyLenssen2019,wang2019dgl,grattarola2021graph,cen2023cogdl,huang2022characterizing}. For example, the popular deep graph library (DGL)~\cite{wang2019dgl} provides rich functionalities to simplify the GNN implementation and dedicated kernels for fast GNN-related computations. However, they rely on CPUs or single GPU only, which is inefficient, especially when it comes to large graphs. Recently, distributed GNN training systems and frameworks~\cite{ma2019neugraph, lin2020pagraph, zhang20212pgraph, liu2023bgl, zheng2020distdgl, zheng2022distributed} have then been developed to enable training GNN models on large graphs by leveraging multiple GPUs or other compute units in a distributed manner. In particular, DistDGL~\cite{zheng2020distdgl, zheng2022distributed} has recently been introduced as an extended package of DGL for distributed GNN training. PyG~\cite{FeyLenssen2019} also supports distributed GNN training recently.

Despite the recent advances, there are still limitations in existing distributed GNN training frameworks and systems. They require the size of the main memory (of a master server) to be much greater than the size of the input graph data. This is because they adopt the conventional \emph{in-memory} graph partitioning algorithms, such as METIS~\cite{karypis1997metis,karypis1998multilevelk}, which require loading the \emph{entire} graph in main memory for partitioning. For example, METIS uses the memory space of up to more than \emph{ten times} the size of the graph data. The feature data is also often loaded together with the graph data in memory for partitioning as in DistDGL and PyG. As real-world graphs continue to grow in size~\cite{sahu2020ubiquity, ying2018graph}, such high memory requirements can be much more problematic, especially with commodity workstations that have only tens of gigabytes of RAM.

In addition, most existing frameworks and systems implicitly assume that the \emph{aggregated} memory size of GPUs (workers) to be much larger than the size of the graph and feature data. As they rely on data parallelism~\cite{li13pytorch,sergeev2018horovod,peng2019generic,jiang2020unified}, they distribute partitioned subgraphs across GPUs for parallel GNN training, where the model is synchronized for each mini-batch via the so-called gradient averaging. Such per-batch synchronization requires each partitioned subgraph (along with its feature data) to fit in the memory of a GPU; otherwise, the partitioned subgraphs need to be loaded and unloaded repeatedly \emph{every} mini-batch. Furthermore, the gradient averaging requires strictly equal partitioning of training samples~\cite{zheng2020distdgl} to ensure \emph{proper} gradient synchronization. Since partitioned subgraphs can hardly be equal-sized, there is a post (random) reassignment of nodes between partitioned subgraphs.

To address these limitations, we propose \n{}, a \emph{memory-efficient} \textbf{s}treaming-based \textbf{d}istributed \textbf{t}raining framework for \textbf{GNN}s. \n{} takes a stream of edges as input, instead of loading the entire graph in main memory, for graph partitioning, in order to reduce the memory space needed for partitioning substantially. We observe that existing streaming partitioning algorithms~\cite{gonzalez2012powergraph, xie2014distributed, petroni2015hdrf, mayer2018adwise, jain2013graphbuilder, mayer2022out}, which are developed for conventional graph mining tasks, \emph{cannot} be directly used for distributed GNN training. This is because they end up fragmenting the neighbors of many nodes across different partitions, which is in conflict with the neighborhood aggregation of GNNs. Thus, we develop \n{} in such a way that the neighborhood aggregation is done properly with the existing streaming partitioning algorithms. Also, to further improve the quality of partitioning, we propose \algoname{}, a novel \textbf{s}treaming \textbf{p}artitioning algorithm based on \textbf{ri}chest \textbf{n}ei\textbf{g}hbors with a linear space complexity for distributed GNN training. It is not only faster than the existing algorithms but also leads to a much smaller number of node replicas for distributed GNN training.

In addition, \n{} supports distributed GNN training even when the aggregated memory size of GPUs is smaller than the size of the graph and feature data. Specifically, \n{} determines and distributes the right amount of workload to each GPU, depending on the available computational resources. That is, \n{} automatically finds the number of partitioned subgraphs, given the number of GPUs, their memory size, graph and feature data size, etc., so that each partitioned subgraph fits in the memory of each GPU without causing an out-of-memory (OOM) error during training. Furthermore, \n{} adopts model averaging instead of gradient averaging for model synchronization across GPUs, where the synchronization is done every epoch or a predefined number of epochs. It is particularly effective when the number of partitions is greater than the number GPUs, in which case the data transfer of partitions no longer need to be done every mini-batch. It also does not need to balance training samples after partitioning.

We summarize our contributions as follows:
\begin{itemize}[itemsep=0pt,leftmargin=1.1em,topsep=0pt]

\item We propose \n{}, a streaming-based framework for distributed GNN training to efficiently scale GNN training to billion-scale graphs under limited memory and computational resources. \n{} also integrates several streaming partitioning algorithms and various GNN models, along with flexible custom modules that allow users to explore new algorithms and models.

\item We develop a novel linear-space streaming partitioning algorithm named \algoname{} as an integral component of \n{}. It improves the performance of graph partitioning by leveraging the richest neighbor information effectively to better capture graph connectivity while being in a streaming fashion.

\item We extensively evaluate the performance of \n{} on seven large public datasets. Experiment results demonstrate the effectiveness and efficiency of \n{}, which greatly reduces the memory footprint (up to $95\%$ less than DistDGL and PyG) without sacrificing the per-epoch training speed and the prediction accuracy. In particular, it is able to handle one of the largest public graph datasets with about $85\%$ less memory footprint than DistDGL and PyG. Furthermore, \algoname{} outperforms state-of-the-art streaming partitioning algorithms significantly in both the running time (up to $5\times$ speed-up) and the replication factor (about 20\% reduction on average).

\end{itemize}

\section{Background and Motivations}
\label{sec:back}

\subsection{Distributed GNN Training}

\begin{figure}[t]
    \vspace{0mm}
    \centering
    \includegraphics[width=0.98\linewidth, trim=0mm 0mm 0mm 0mm, clip]{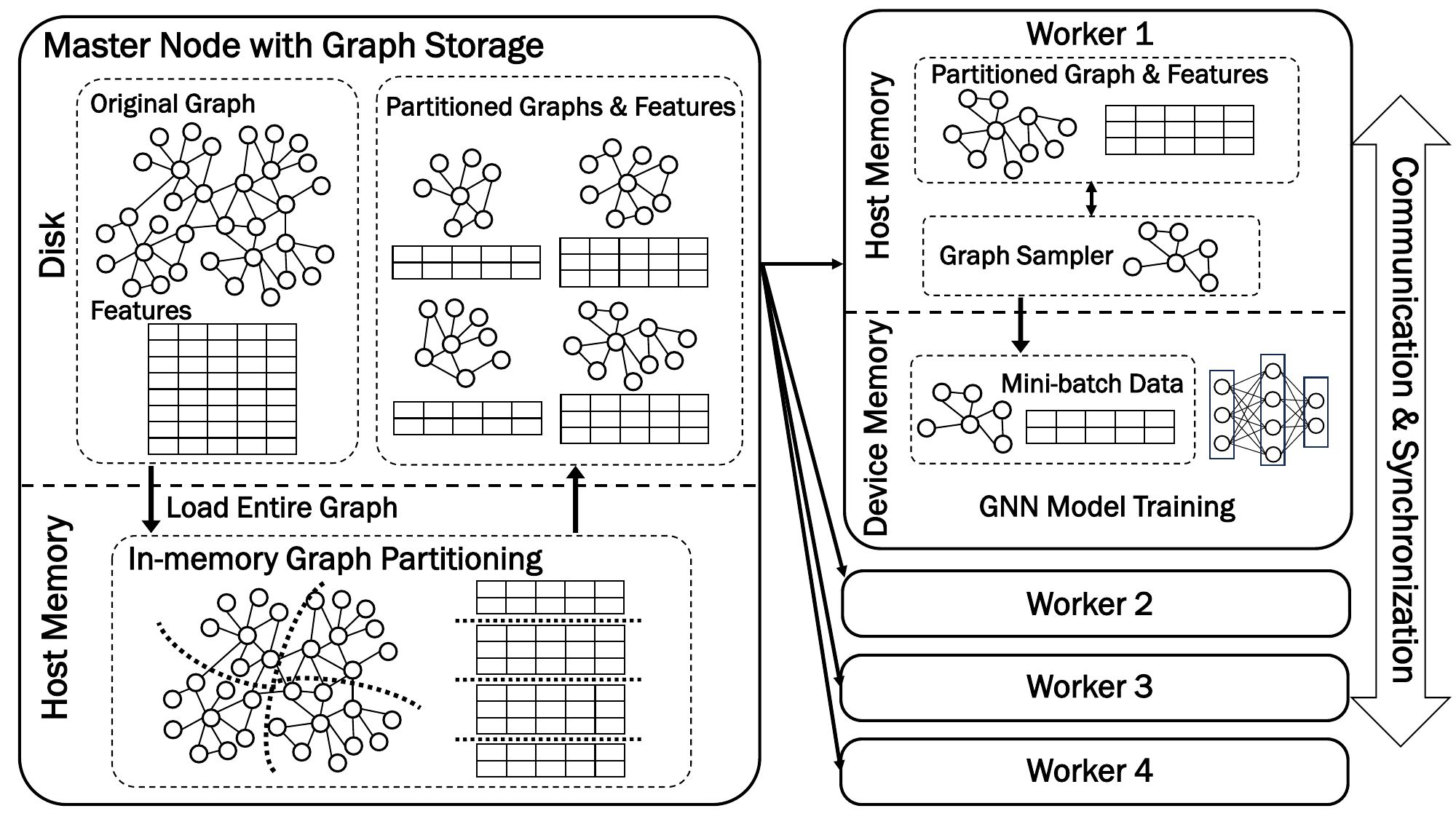}
    \vspace{-2mm}
    \caption{End-to-end workflow of a distributed GNN training system.}
    \vspace{-3mm}
    \label{fig:dist}
\end{figure}

Distributed GNN training frameworks and systems~\cite{ma2019neugraph, lin2020pagraph, zhang20212pgraph, liu2023bgl, zheng2020distdgl, zheng2022distributed} have been developed to improve the training speed as well as the scalability in building GNN models on large graphs. As shown in Figure~\ref{fig:dist}, the general process for distributed GNN training is as follows.

An \emph{entire} input graph and its associated node features are first loaded to the main memory of a master server. They are then partitioned by the master server using some \emph{in-memory} graph partitioning algorithm, e.g., METIS~\cite{karypis1998multilevelk}. The resulting subgraphs along with their corresponding node features are distributed to different workers, e.g., GPUs or compute units, in the cluster. Each worker then processes its assigned subgraph for GNN training, which involves the neighborhood aggregation as a key operation~\cite{ma2021deep, hamilton2020graph}.

For training, the workers need to communicate with each other to obtain the necessary information, including the full set of neighbors for each target training node. The training process also generally involves a graph sampling process, especially when the model is trained in a mini-batch manner~\cite{hamilton2017inductive,zheng2020distdgl}. In addition, to update model weights, gradient synchronization~\cite{li13pytorch,sergeev2018horovod} is done among all workers upon completion of their computation of gradients for \emph{each mini-batch} of samples. 

\begin{table}[t]
\renewcommand{\arraystretch}{1.2}
    \vspace{0mm}
    \caption{Data size vs. peak memory usage of METIS in GB}
    \vspace{-2mm}
    \label{table:memory}
    \centering
    \footnotesize
    \begin{adjustbox}{width=0.99\columnwidth,center}
        \begin{tabular}{|c|c|c|c|c|c|c|}
        \hline
        & OGB-Arxiv & Yelp & OGB-Products & Reddit & Amazon & OGB-Papers \\
        \hline
        \hline
        Raw graph data size & 0.03 & 0.18 & 1.68 & 1.39 & 3.59 & 54.09 \\
        \hline
        Peak memory of METIS & 0.81 & 6.77 & 21.19 & 4.90 & 48.64 & \textcolor{red}{OOM} \\
        \hline
        \end{tabular}
    \end{adjustbox}
\vspace{3mm}
\end{table}

\subsection{Limitations in Distributed GNN Training}

We below uncover several limitations in existing distributed GNN training frameworks/systems and when using existing streaming-based partitioning algorithms for distributed GNN training. 

\vspace{1pt}
\noindent \textbf{Drawbacks with in-memory partitioning.} Graph partitioning is the first step in distributed GNN training, so the quality of the partitions has a significant impact on the performance of downstream tasks in terms of memory footprint, communication cost, and evaluation accuracy. To our knowledge, all the existing distributed GNN training frameworks~\cite{ma2019neugraph, lin2020pagraph, zhang20212pgraph, liu2023bgl, zheng2020distdgl, zheng2022distributed} require loading the \emph{entire} graph \emph{in main memory} and leverage the conventional partitioning algorithms (e.g., METIS~\cite{karypis1998multilevelk}) for partitioning. However, the rapidly growing size of current graphs~\cite{sahu2020ubiquity} can easily exceed the main memory size of most commodity workstations or servers. 

Table~\ref{table:memory} shows the peak memory usage of METIS algorithm used in DistDGL~\cite{zheng2020distdgl, zheng2022distributed}. METIS requires the memory space of more than ten times the size of raw data to partition a graph. In particular, it fails with OOM errors in a commodity server with 96~GB RAM when partitioning the OGB-Papers dataset~\cite{hu2020open}. This is one of the largest datasets for which DistDGL requires about 400~GB RAM~\cite{dgl_metis}.\footnote{Note that DistDGL and PyG load the graph data along with the feature data for partitioning. Thus, their actual memory usage is greater than that of METIS. In addition, DistDGL also supports ParMETIS, a parallel version of METIS for graph partitioning using multiple machines. However, it still requires loading the entire graph into memory, leading to the high memory usage and potential OOM errors~\cite{dgl_parmetis}.}  Therefore, to deal with large graphs using \emph{in-memory} partitioning algorithms, the only viable solution is deemed to update the hardware with enough RAM space, but it is \emph{not} a cost-efficient and scalable solution.

\begin{figure}[t]
\captionsetup[subfloat]{captionskip=2pt}
\centering
    \vspace{0mm}
    \subfloat[GAT]{%
        \includegraphics[width=0.45\linewidth, trim=0cm 0cm 0cm 0cm, clip]{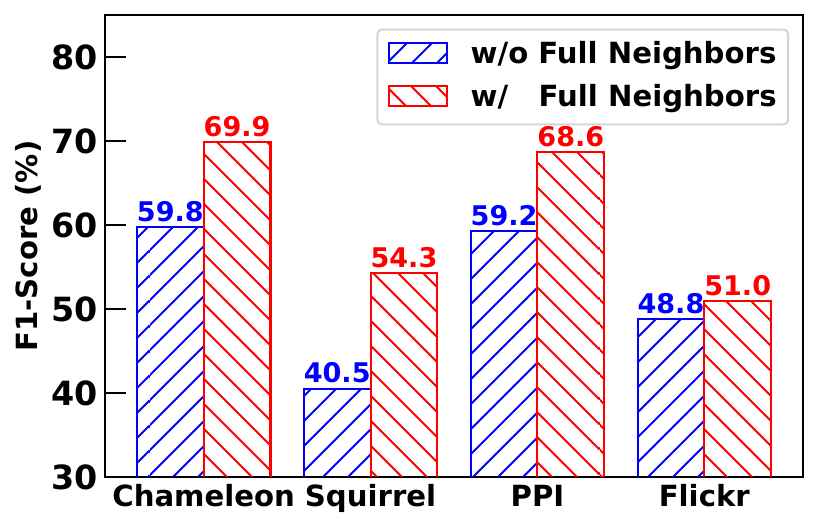}
    }
    \hspace{3mm}
    \subfloat[GraphSAGE]{%
        \includegraphics[width=0.45\linewidth, trim=0cm 0cm 0cm 0cm, clip]{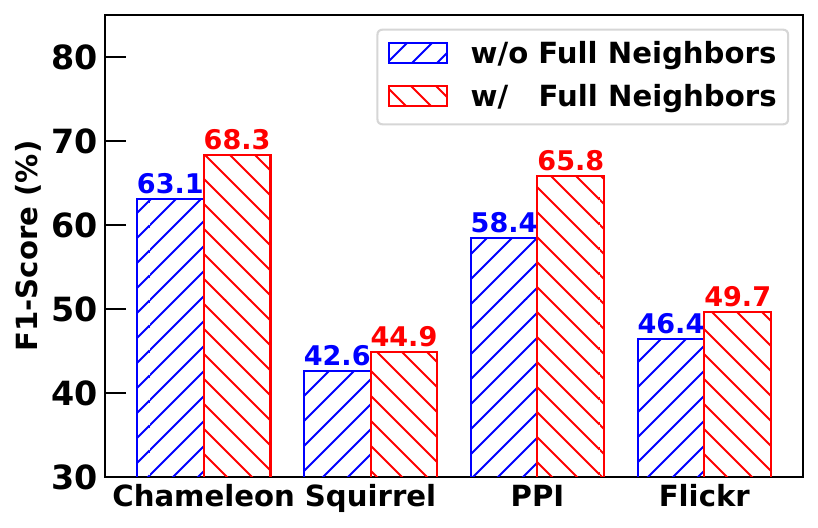}
    }
    \vspace{-1mm}
    % \Description{}
    \caption{Node classification accuracy of (a) GAT and (b) GraphSAGE trained on partitioned subgraphs by 2PS-L with (w/) and without (w/o) full neighbors added to the subgraphs.}
    \label{fig:acc_drop}
    \vspace{0mm}
\end{figure}

\vspace{1pt}
\noindent \textbf{Ineffectiveness of existing streaming-based partitioning algorithms.} There have been several streaming-based partitioning algorithms~\cite{gonzalez2012powergraph, xie2014distributed, petroni2015hdrf, mayer2018adwise, jain2013graphbuilder, mayer2022out} to deal with large graphs, albeit not proposed for distributed GNN training. Instead of loading the entire graph in memory, they take in a stream of edges as input and use partial knowledge of the graph to make partitioning decisions \emph{on the fly}. They often partition the graph into roughly equal-sized subgraphs by mostly cutting through \emph{high-degree} nodes, i.e., replicating high-degree nodes across partitions, making the full neighbors of each node fragmented across partitions. Such partitioning, however, becomes \emph{ineffective} when used for distributed GNN training, as most GNNs require the full neighborhood information for feature aggregation~\cite{lin2020pagraph, zhang20212pgraph, liu2023bgl, zheng2020distdgl,zhang2020agl,md2021distgnn}. 

Figure~\ref{fig:acc_drop} shows the node classification accuracy of GAT and GraphSAGE trained on graphs, which are partitioned by the recent edge streaming-based algorithm 2PS-L~\cite{mayer2022out}. We observe that the performance is \emph{unsatisfactory} when they are trained on the partitioned subgraphs \emph{without} having the full neighborhood information. By including the full neighbor list of each node, the performance becomes almost identical to the one achieved when they are trained centrally. However, the resulting replication factor can be doubled. Thus, it calls for an effective streaming-based partitioning algorithm for distributed GNN training, which partitions the graph into roughly equal-sized components with a low replication factor while maintaining the full neighbor list of each node in a partition.

\vspace{1pt}
\noindent \textbf{Inflexibility of gradient averaging for model synchronization.} Distributed GNN training frameworks are developed mostly based on data parallelism, where each worker trains the model on its assigned partitioned subgraph, and periodic model synchronization, where gradient averaging is done across the workers for \emph{each mini-batch}. Thus, they require the number of mini-batches, i.e., the number of training nodes in each subgraph, to be exactly the same for all the workers. If there is a mismatch in the numbers of mini-batches among the workers, the training would halt, since in the last few batches, the worker with the least number of mini-batches would have no gradient to share with the other workers for synchronization~\cite{li13pytorch}. To resolve this issue, for example, DistDGL~\cite{zheng2020distdgl, zheng2022distributed} proposes to (randomly) move nodes from larger partitions to smaller ones on top of the partitioning results obtained by METIS. However, such a reassignment of nodes between partitions is ineffective, as the connectivity of the graph is ignored. 

\begin{figure}[t]
    \vspace{0mm}
    \centering
    \includegraphics[width=0.92\linewidth, trim=0mm 0mm 0mm 0mm, clip]{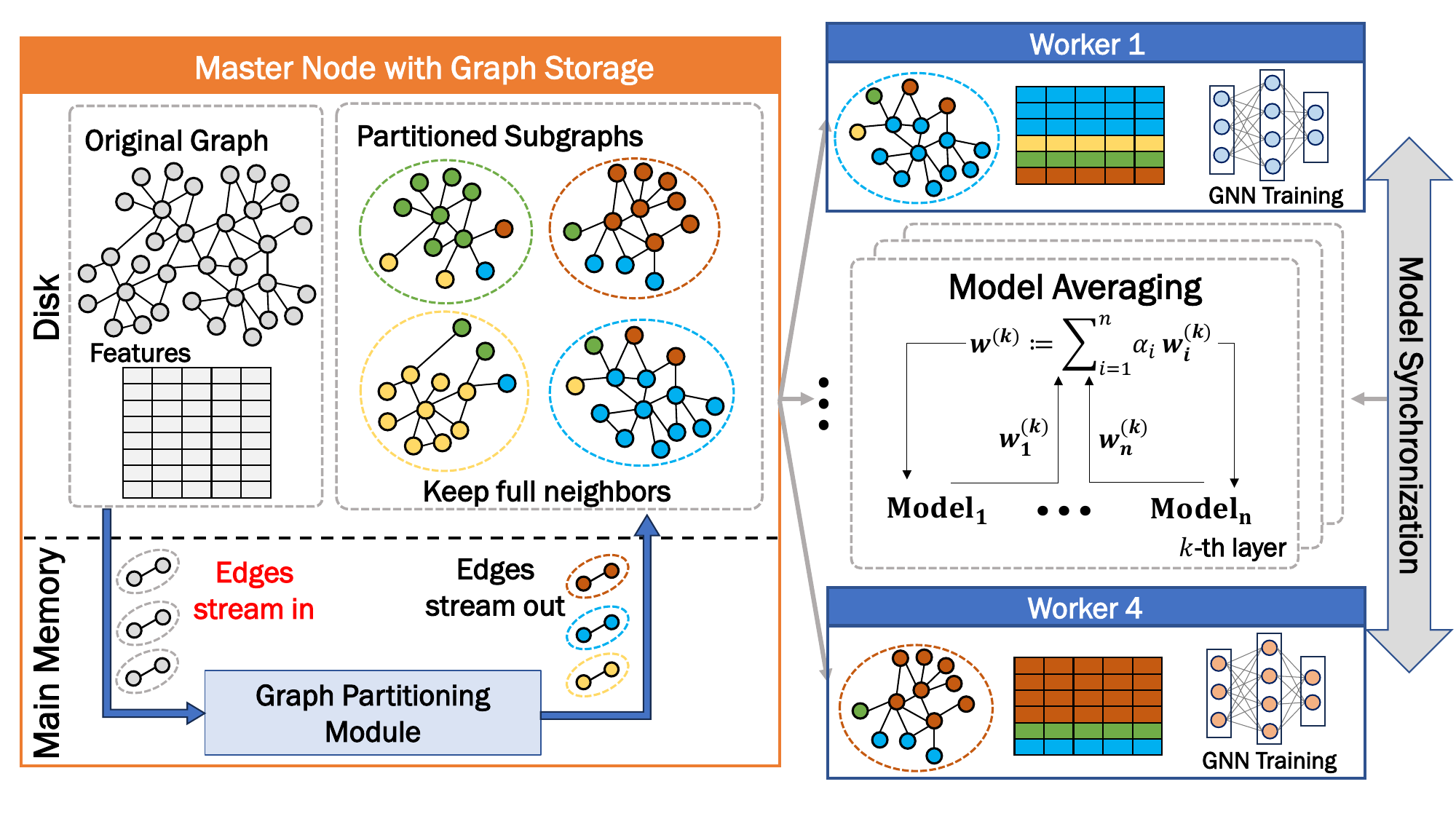}
    \vspace{-3mm}
    \caption{Overview of \n{}.}
    \vspace{0mm}
    \label{fig:system}
\end{figure}

In addition, the existing distributed GNN training frameworks and systems implicitly assume that each partitioned subgraph (along with its feature data) can fit in the memory of each worker. If it is otherwise, the partitioned subgraphs need to be further partitioned so that each worker can handle a partition. In that case, however, the gradient averaging requires each worker to load and unload multiple partitions in sequence for computing its gradient \emph{every} mini-batch, which incurs a high data transfer overhead and, in turn, increases the overall training time. Therefore, there is a need for a more flexible and efficient synchronization strategy to boost the efficiency of distributed GNN training.

\section{\n{}}\label{sec:sys}

We propose \n{}, a streaming-based distributed training framework for GNNs, which effectively addresses the aforementioned current limitations in distributed GNN training to enable training GNN models under limited computational resources.

\subsection{Overview}

The overall architecture of \n{} is shown in Figure~\ref{fig:system}. \n{} employs a data-parallelization approach to distribute both data and computation across workers for training GNN models. Different from the existing GNN training frameworks, \n{} takes in a \emph{stream of edges} as input and uses our novel streaming-based partitioning algorithm named \algoname{} for partitioning (Section~\ref{sec:sys-partition}). \n{} allocates the right amount of workload to each worker for model training in parallel, depending on the available computational resources (Section~\ref{sec:sys-workload}). Unlike the existing training frameworks that perform the gradient averaging across the workers every mini-batch for model synchronization, in \n{}, we propose to use model averaging for model synchronization (Section~\ref{sec:sys-gnn}).

\subsection{Graph Partitioning}\label{sec:sys-partition}

To train GNN models on a large graph in a distributed manner, we need to partition the graph into roughly equal-sized subgraphs so that they can be distributed to the available workers for parallel computing and load balancing. The partitioned subgraphs also need to maintain the neighborhood information of each node for the neighborhood aggregation in GNNs. To this end, we propose \algoname{} as a main streaming-based partitioning algorithm in \n{} for distributed GNN training.

\setlength{\algomargin}{0.7em}
\begin{algorithm2e}[t!]
\small
\SetAlgoNoEnd
\setstretch{1.13}
\caption{\algoname{}}\label{alg:\algoname{}}
\KwIn{the number of partitions $p$; balance factor $\beta$}
\KwOut{partitions $\mathcal{P}_1, \mathcal{P}_2, \ldots, \mathcal{P}_p$}
{$k \leftarrow 1;$}\\
\tcc{Node clustering}
\For {{\bf{each}} \emph{edge} $(u,v) \in E$}{
    \If{$c_u=0$}{
        $c_u \leftarrow k$;~ $k \leftarrow k+1$;\\ 
        $vol(c_u) \leftarrow vol(c_u)+d(u)$;~ $C_{c_u} \leftarrow \{u\}$; }
    \If{$c_v=0$}{
        $c_v \leftarrow k$;~ $k \leftarrow k+1$;\\ 
        $vol(c_v) \leftarrow vol(c_v)+d(v)$;~ $C_{c_v} \leftarrow \{v\}$; }
    \If{$c_u \neq c_v$ {\bf{and}} $vol(c_u) \leq \tau_{vol}$ {\bf{and}} $vol(c_v) \leq \tau_{vol}$}{
        \If{$vol(c_u) \leq vol(c_v)$}
        {Move $u$ from $C_{c_u}$ to $C_{c_v}$ and\\ update $vol(c_u), vol(c_v)$;}
        \Else{
        Move $v$ from $C_{c_v}$ to $C_{c_u}$ and\\ update $vol(c_u), vol(c_v)$;}}
        \lIf{ $d(n_u) < d(v)$ }{
            $n_u\leftarrow v$
        }
        \lIf{ $d(n_v) < d(u)$ }{
            $n_v\leftarrow u$
        }
        }
\tcc{Cluster merging}
\For {{\bf{each}} \emph{node} $v \in V$}{
         \If{$d(n_{r(c_v)}) < d(n_v)$}{
                $r(c_v) \leftarrow v$; 
        }
}
{Sort clusters by size in ascending order;} \\
\For {{\bf{each}} \emph{non-empty cluster} $i$}{
    $t \leftarrow n_{r(i)}$; \\
    \If{$|C_i| + |C_{c_t}| \leq \beta \cdot |V|/p$ \bf{and} $i \neq c_t$}{
        $C_{c_t} \leftarrow C_{c_t} \cup C_i$;~ $C_i \leftarrow \emptyset$;}}
\tcc{Assigning clusters to partitions}
{Sort clusters by size in descending order;} \\
\For {{\bf{each}} \emph{non-empty cluster} $i$}{
    $t \leftarrow \argmin_s |\mathcal{P}_s|$; \\
    $\mathcal{P}_t \leftarrow \mathcal{P}_t \cup C_i$;} 
\end{algorithm2e}

For an undirected graph $\mathcal{G} \!=\! (V, E)$, where $V$ is the set of nodes, and $E \!\subseteq\! V \!\times\! V$ is the set of edges, we first collect the following definitions.\footnote{Since an undirected edge is represented by two directed edges in opposite directions, our algorithm can be easily extended to directed graphs.} Let $d(v)$ be the degree of node $v$, and let $c_v$ be the index of the cluster that node $v$ belongs to. Let $C_{i}$ be the set of nodes in cluster $i$, and let $vol(i)$ be the volume of cluster $i$, i.e., $vol(i) \!=\! \sum_{v \in C_i} d(v)$. We below explain the operations of \algoname{}, as depicted in Algorithm~\ref{alg:\algoname{}}. It has three main steps, which are node clustering, cluster merging, and assigning clusters to partitions. Note that the node clustering step is done in a streaming manner, but the other steps are not streaming operations.

\vspace{1pt}
\noindent \textbf{Node clustering (Lines 2--17).} We adopt the streaming clustering algorithm proposed in~\cite{hollocou2017streaming} for node clustering as a first step of \algoname{}. It aims to uncover natural clusters of a graph but has \emph{no control} over the number of clusters identified.  Specifically, for each edge $(u, v)$ that streams in, if one of its end nodes has not been seen before, we assign it to a new cluster and use its degree to update the volume of the new cluster (\textbf{Lines 3--8}). The node degree information is here assumed to be available, but it can be easily obtained or estimated online from the edge stream~\cite{xie2014distributed, petroni2015hdrf, mayer2022out}. Next, if $u$ and $v$ belong to two different clusters, and if the volumes of the two clusters are both smaller than a given threshold $\tau_{vol}$, we move the node that is in the lower volume cluster to the higher volume cluster and update the cluster volumes by adding or subtracting the node degree accordingly (\textbf{Lines 9--15}). Otherwise, we continue to process the next edge. Note that we also record the richest neighbor $n_v$ of each node $v$, i.e., the largest degree node among its neighbors, which continues to be updated as edges are streamed in (\textbf{Lines 16--17}) and will be used in the next step.

\vspace{1pt}
\noindent \textbf{Cluster merging (Lines 18--25).} While the above clustering algorithm is simple and effective, we observe that it returns a significant number of small (disjoint) clusters, as shown in Table~\ref{table:cluster}. Many of them could be part of bigger clusters.  We thus devise an effective cluster-merging method to merge the clusters, identified by the clustering algorithm, into \emph{fewer} clusters. The key idea behind this method is to use the richest neighbor of each node (the largest degree node among its neighbors) to identify a possible connection between two clusters. Intuitively, high-degree nodes are more likely to appear in large clusters. Note that the preferential attachment in the Barab\'{a}si-Albert model~\cite{albert2002statistical} makes the rich (high-degree nodes) get richer (having higher degrees) and in turn gives rise to a power-law degree distribution, which is commonly observed in the datasets. Hence, we propose to use the richest neighbor of each node in a cluster to merge the cluster into a potentially larger one where the richest neighbor belongs.

\begin{table}[t]
\renewcommand{\arraystretch}{1.2}
\vspace{-1mm}
\caption{Results of the clustering algorithm~\cite{hollocou2017streaming} vs. \algoname{}}
\vspace{-3mm}
\label{table:cluster}
\centering
\footnotesize
\begin{adjustbox}{width=\columnwidth,center}
    \begin{tabular}{|c|c|c|c|c|c|c|c|}
        \hline
        & & Flickr & OGB-Arxiv & Yelp & OGB-Products & Reddit & Amazon \\
        \hline
        \hline
        \# of & Clustering &  10201 & 24544 & 33256 & 52067 & 12266 & 311244 \\
        \cline{2-8}
        clusters & \algoname{} & 140 & 250 & 6345 & 31 & 22 & 85610 \\
        \hline
        \hline
        Avg.  & Clustering & 9$\pm$62 & 7$\pm$77 & 18$\pm$261 & 41$\pm$1488 & 19$\pm$463 & 4$\pm$650 \\
        \cline{2-8}
        cluster size & \algoname{} & 638$\pm$518 & 677$\pm$1085 & 113$\pm$2156 & 77439$\pm$107723 & 10590$\pm$12506 & 18$\pm$2321 \\
        \hline
    \end{tabular}
\end{adjustbox}
\vspace{-0mm}
\end{table}

Specifically, for each cluster $i$, we first identify its `representative node', say, $r(i)$, which is defined as the node whose richest neighbor is the largest (the top richest neighbor) among the richest neighbors of the nodes in cluster $i$ (\textbf{Lines 18--20}).\footnote{If there is a tie, it breaks randomly.} We then sort the clusters based on their cluster size, which is defined as the number of nodes in the cluster, in ascending order. Starting from the smallest cluster, we merge each cluster into the cluster that the richest neighbor of its representative node belongs to, if the target cluster is a different one, and if the number of nodes in the merged cluster is going to be smaller than a given threshold (\textbf{Lines 22--25}). In other words, letting $i$ and $j$ be two different clusters for a merger, we merge them if $|C_{i}| + |C_{j}| \leq \beta \cdot |V|/p$, where $p$ is the given number of partitions, and $\beta$ is the balance factor. We set $\beta\!=\!1.05$ in our experiments. Note that the representative node and its richest neighbor can belong to the same cluster, in which case nothing happens. Once the two clusters are merged, we update the representative node of the merged cluster and place this cluster at the correct rank based on its merged size, i.e., the number of nodes in the merged cluster. The entire merging step is repeated until all clusters are scanned.

We observe that the above cluster-merging method can reduce the number of clusters up to \emph{three orders of magnitude}, as shown in Table~\ref{table:cluster}. We here emphasize that our cluster-merging method does not merge clusters randomly but in a systematic way by identifying the connections between them based only on the richest neighbor of the representative node of each cluster.

\begin{figure}[t]
    \vspace{-1mm}
    \centering
    \includegraphics[width=0.9\linewidth, trim=0mm 0mm 0mm 0mm, clip]{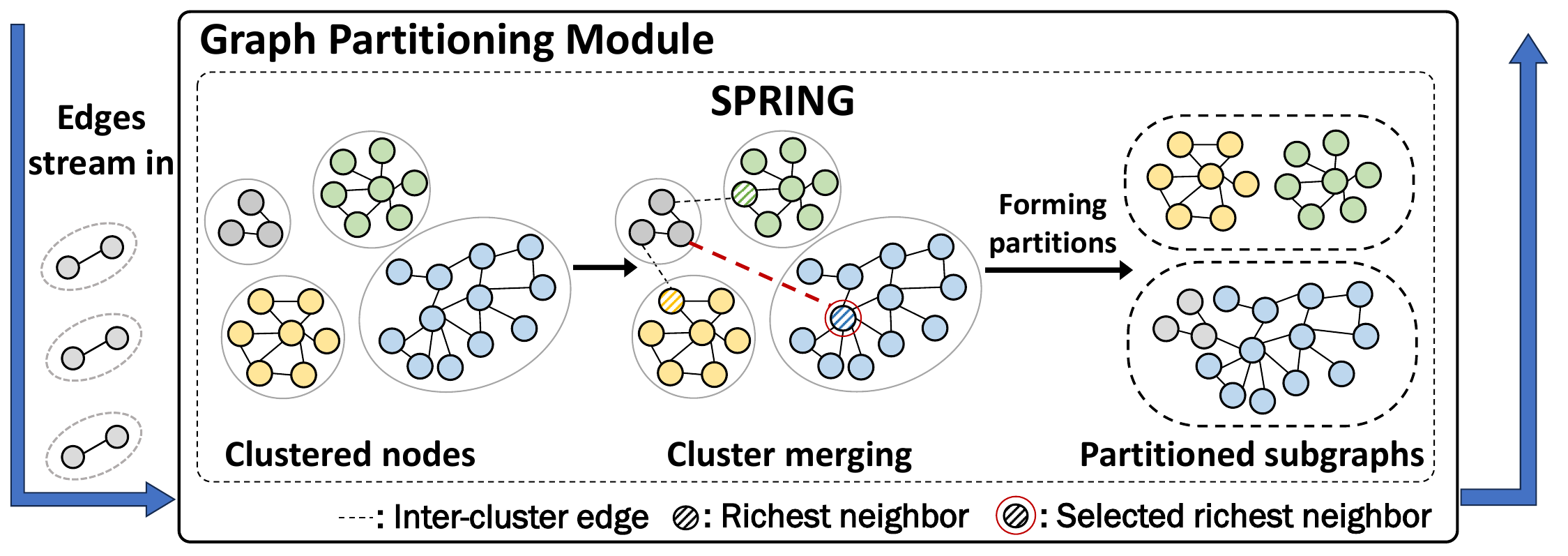}
    \vspace{-3mm}
    \caption{Illustration of \algoname{} as a part of \n{}.}
    \vspace{0mm}
    \label{fig:connect_rich}
\end{figure}

\vspace{1pt}
\noindent \textbf{Assigning clusters to partitions (Lines 26--29).} Given the merged clusters, we assign them to $p$ partitions such that the resulting partitions are of roughly equal size. This problem can be viewed as the classical `minimum makespan scheduling' problem, which is NP-hard. The greedy sorted list scheduling algorithm is widely known as a simple yet effective approximation algorithm for the problem~\cite{graham1969bounds,leung2004handbook}. Specifically, we sort the clusters in descending order in terms of their cluster size and assign them one by one to the currently least occupied partition. \algoname{} finally returns $p$ disjoint partitions. That is, each node belongs to a unique partition. This is the end of partitioning $|V|$ nodes into $p$ partitions.

What remains to be determined is which partition each edge, especially the one spanning different partitions, should be assigned to. Thus, we take a stream of edges again to decide which partition each edge is assigned to. For each edge, if its two end nodes belong to the same partition, then it naturally goes to the partition. However, if they are assigned to different partitions, such a `cross-partition' edge is replicated and added to both partitions. This way the full neighbor list of each node is preserved in its corresponding partition. Note that the resulting node replicas are not considered as target nodes for GNN training and testing. See Figure~\ref{fig:connect_rich} for an illustration on how \algoname{} works overall.

\begin{figure*}[t]
\captionsetup[subfloat]{captionskip=2pt}
\centering
    \vspace{0mm}
    \subfloat[The number of partitions is equal to the number of workers]{%
        \includegraphics[width=0.38\linewidth, trim=0cm 0cm 0cm 0cm, clip]{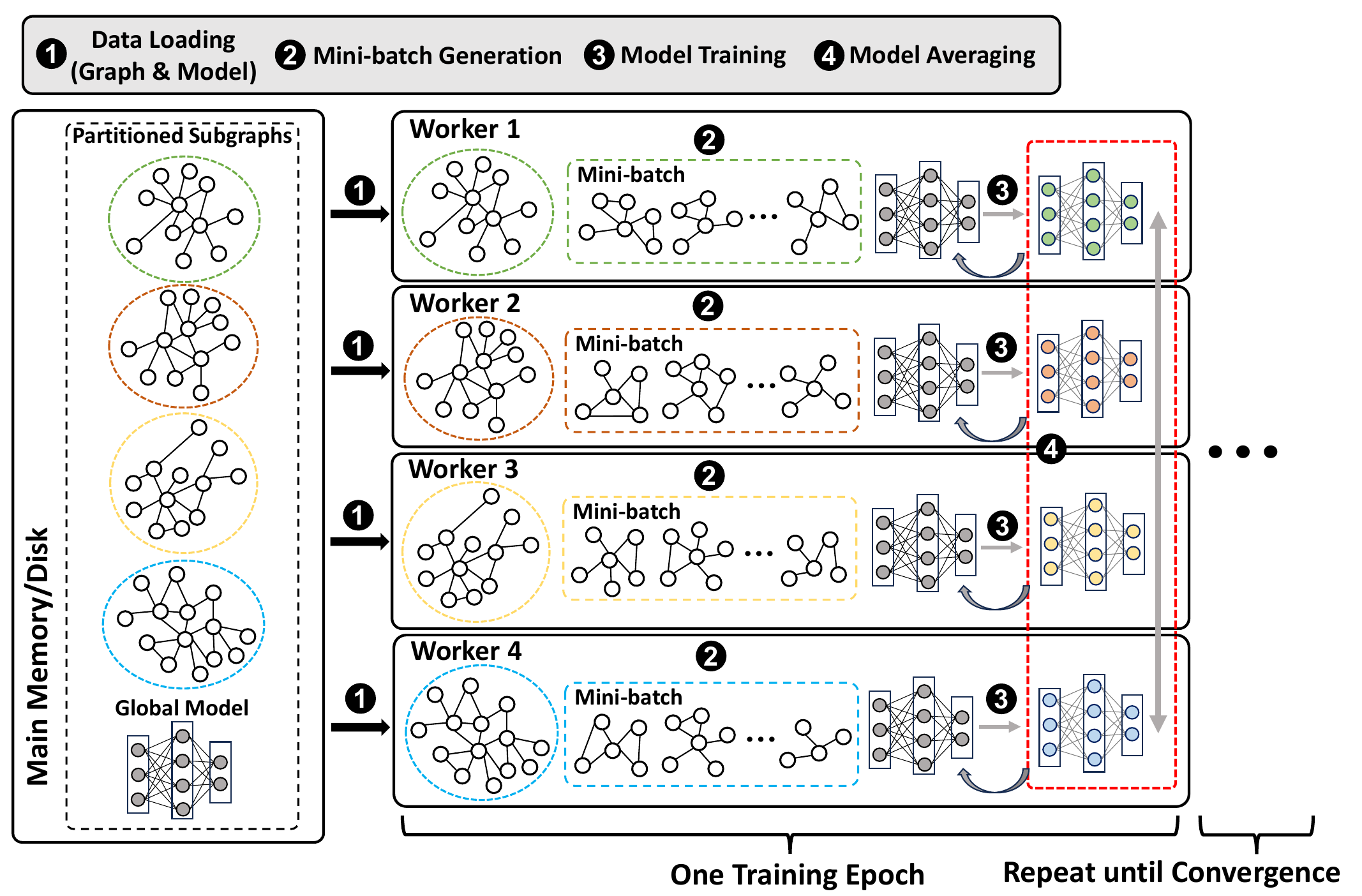}
    }
    \hspace{3mm}
    \subfloat[The number of partitions is greater than the number of workers]{%
        \includegraphics[width=0.55\linewidth, trim=0cm 0cm 0cm 0cm, clip]{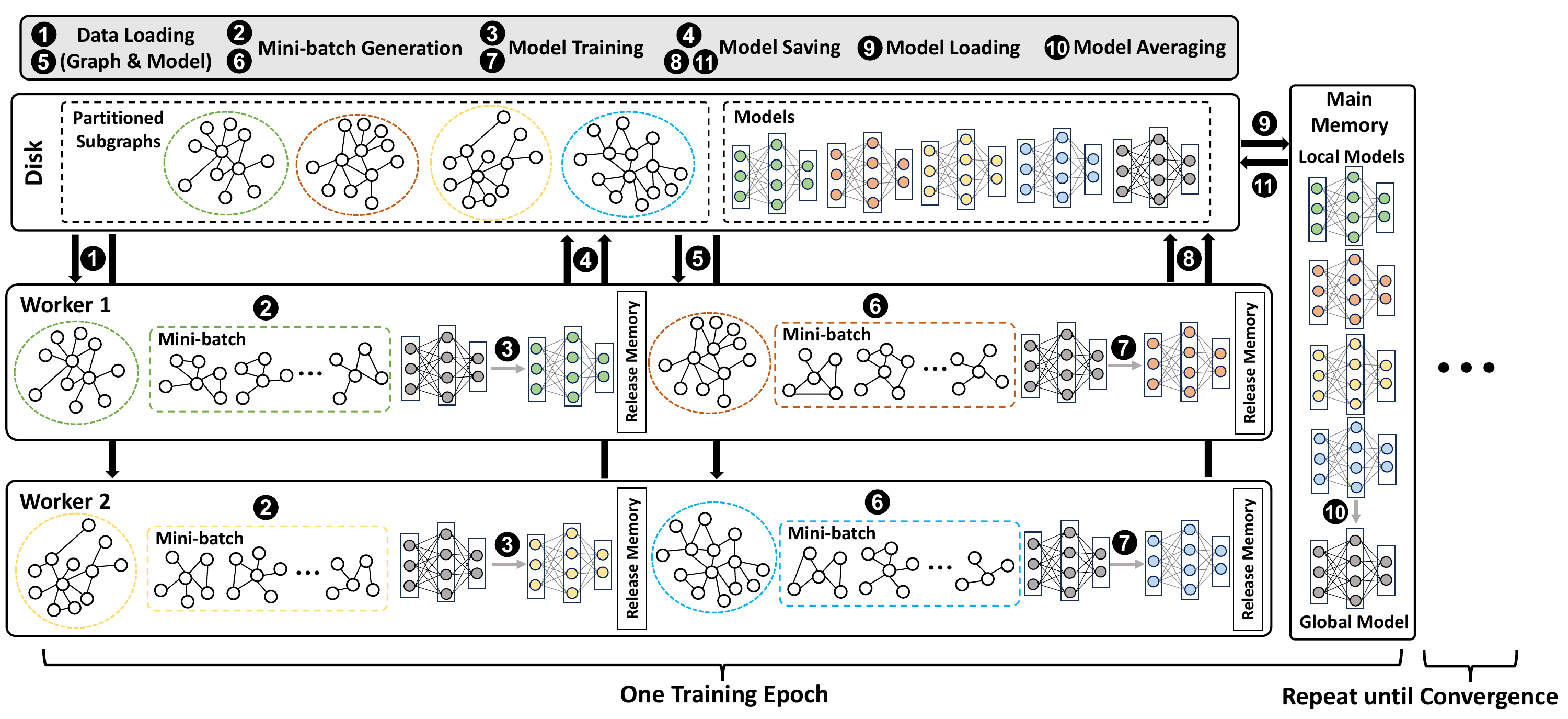}
    }
    \vspace{0mm}
    \caption{Training workflow of \n{} after the input graph is partitioned.}
    \label{fig:workflow}
    \vspace{-3mm}
\end{figure*}

\vspace{1pt}
\noindent \textbf{Node-feature file partitioning.} For a large graph, its associated node-feature file can also hardly be handled by a master server with limited memory. To address this issue, we propose to use the memory-mapped file (\texttt{mmap})~\cite{love2013linux} to efficiently split the feature file. The \texttt{mmap} is used to access and manipulate data stored in a large file on disk \emph{without} the need to load the entire file into RAM. It creates a mapping between the file on the disk and the memory in the program's address space and is extremely efficient for random access of the file, which meets our requirement for splitting the large node-feature file. Thus, we create an \texttt{mmap} for the feature file. Once partitions are identified, in \n{}, we go over the nodes in each partition to get their corresponding node features using the \texttt{mmap} and save them into partitioned feature files.

\vspace{1pt}
\noindent \textbf{Complexity analysis of \algoname{}.} We formally analyze the time and space complexity of \algoname{} as follows:

\begin{theorem}\label{complexity1}
\algoname{} has a time complexity of $\mathcal{O}(|E|+|V|\log|V|)$ and a space complexity of $\mathcal{O}(|V|)$.
\end{theorem}
% \vspace{-3mm}
\begin{proof}
\noindent \textbf{Time complexity.}
In \algoname{}, the time complexity of the node clustering step is linear to the number of edges, i.e., $\mathcal{O}(|E|)$, as we need to take a stream of edges and perform constant-time operations on each edge. In the cluster merging step, it requires $\mathcal{O}(|V|)$ to find the representative node of each cluster, as we need to check each node in the cluster. It then takes $\mathcal{O}(|V|\log|V|)$ to sort the clusters by size (\textbf{Line 21}), since every cluster can contain only one node in the worst case. In practice, it should be much faster, as the number of clusters is usually much smaller than $|V|$. As shown in Tables~\ref{table:dataset} and~\ref{table:cluster}, the difference can be several orders of magnitude. 

In addition, observe that merging two clusters can be done in constant time (e.g., using doubly linked lists). Also, reordering a newly merged cluster to place it at the correct rank requires $\mathcal{O}(\log|V|)$ (e.g., via binary search), as only the size of the merged cluster has changed. Since these operations can be repeated up to $\mathcal{O}(|V|)$ times, it takes $\mathcal{O}(|V|\log|V|)$ in total. Once the clusters are obtained, we next assign each cluster to the currently least occupied partition. This whole process takes $\mathcal{O}(|V|\log p)$, as we keep the $p$ partitions sorted by their size and make each assignment accordingly. Finally, we take the edge stream again, which takes $\mathcal{O}(|E|)$, to ensure that the full neighbor list of each node is maintained in its partition. Therefore, \algoname{} has a time complexity of $\mathcal{O}(|E|+|V|\log|V|)$. 

\vspace{1pt}
\noindent \textbf{Space complexity.}
For the node clustering step, we need to store node degrees, cluster volumes, and the set of nodes in each cluster. Each of these has a space complexity of $\mathcal{O}(|V|)$. For the other steps, we need to store cluster sizes, the richest neighbor of each node, the representative node of each cluster, and the partition assignment of each node. Each of these has a space complexity of $\mathcal{O}(|V|)$. Thus, \algoname{} has a space complexity of $\mathcal{O}(|V|)$.
\end{proof}

\subsection{Workload Allocation}\label{sec:sys-workload}

\n{} is designed in a way that allocates and distributes the right amount of workload to each worker (GPU). \n{} first automatically determines the number of partitions according to the available resources. Specifically, if the cluster has $q$ GPUs with each having a memory size of $M$~GB, and the size of graph and feature data, say, $S$~GB, is smaller than $q (M\!-\! T)$~GB, i.e., $S \!\leq\! q (M\!-\! T)$, then the number of partitions is $q$, where $T$ is the amount of memory used for GNN-related computations on a single GPU ($T \!<\! M$). If $S \!>\! q (M\!-\!T)$, the graph is partitioned into $\ceil{S/(M-T)}$ subgraphs of (roughly) equal size. For the former case, each GPU only needs to process a single partitioned subgraph for training, while for the latter case, each GPU processes a few subgraphs in a sequential order. \n{} also allows users to specify the number of partitions as a parameter in case they have a desired number of partitions. 

In \n{}, we use a conservative estimate of $T$ to avoid the possible OOM problem during model training. Specifically, we train GNN models on several small-graph datasets using a single GPU and profile their memory usages on both CPU and GPU via the `pytorch\_memlab' profiler~\cite{pytorch_memlab}. We first observe that the memory usage of the graph object is more or less the same as the size of the graph and feature data stored on the disk. Thus, we just use the data size on the disk as an estimate of their size in memory. We also observe that the \emph{total} GPU memory usage of training a 3-layer GAT model is often about two to three times the size of the graph object in memory. Note that the former includes the graph-object size, and the GAT model is the most computationally expensive one among the GNN models integrated into \n{}. Thus, by default, we use two-thirds of the total available memory space of a single GPU as a conservative estimate of $T$. Nonetheless, \n{} allows users to specify the value of $T$ as a parameter.

Given the number of partitions and the number of available GPUs, \n{} then allocates the training workload as follows. When the number of partitions, say, $p$, is smaller than or equal to the number of available GPUs, say, $q$, the first $p$ GPUs naturally have one partitioned subgraph each for model training. See Figure~\ref{fig:workflow}(a) for the workflow in this case. For the case that $p \!>\! q$, we cyclically assign $p$ subgraphs to each GPU. Each GPU handles $\floor{p/q}$ or $\ceil{p/q}$ subgraphs and trains $\floor{p/q}$ or $\ceil{p/q}$ local models independently and sequentially. Once all $p$ local models are trained, they are synchronized and updated via model averaging, which will be explained shortly. After that, each GPU continues to train based on the updated model replica. This process is repeated until convergence or for a predefined maximum number of training epochs. See Figure~\ref{fig:workflow}(b) for an illustration of its training workflow. This design enables \n{} to train GNN models on \emph{large} graphs under limited computational resources, thereby making \n{} cost-efficient and scalable, as shall be demonstrated in Section~\ref{sec:distdgl}.

\subsection{Distributed Model Training}\label{sec:sys-gnn}

As mentioned in Section~\ref{sec:back}, the gradient averaging for model synchronization is inflexible and inefficient, especially when training GNN models on \emph{large} graphs under limited resources. Thus, in \n{}, we propose to use model averaging~\cite{li2019convergence,stichlocal,yu2019parallel} for model synchronization. Each worker trains its local view of the model on its assigned data for each epoch or a predefined number of epochs. The model weights are then averaged and synchronized among the workers. Specifically, as shown in Figure~\ref{fig:system}, worker $i$ obtains its model weights, say, $\bm{w}_i^{(1)}, \bm{w}_i^{(2)}, \ldots, \bm{w}_i^{(k)}$, upon its assigned subgraph, where $\bm{w}_i^{(k)}$ is the model weights at the $k$-th layer. Note that the assigned subgraph here can be a collection of a few partitioned subgraphs. The model weights from different workers are then averaged proportionally to the numbers of their training nodes, and the averaged model weights, say, $\bm{w}^{(1)}, \bm{w}^{(2)}, \ldots, \bm{w}^{(k)}$, are synchronized among the workers, where $\bm{w}^{(j)} \!:=\! \sum_i \alpha_i \bm{w}_i^{(j)}$, $j = 1, \ldots, k$, and $\alpha_i$ is the ratio of the number of training nodes in worker $i$ to the total number of training nodes. 

Model averaging, also known as local SGD and federated averaging, is not a new technique~\cite{li2019convergence,stichlocal,yu2019parallel}. Most studies have, however, examined under what conditions the model averaging achieves the same linear speedup as parallel SGD (gradient averaging) does with respect to the number of workers. In contrast, in \n{}, we leverage model averaging as an effective synchronization strategy to enable GNN training even when the size of graph data is greater than the aggregated memory size, in which case each worker needs to repeatedly load and unload different sets of local data samples (partitioned subgraphs). In other words, the benefit of model averaging here is more on the cost of data transfer rather than the synchronization cost. We will demonstrate in Section~\ref{sec:eval} that the convergence speed and model performance of GNN models in \n{} are almost identical to the ones with gradient averaging. Note that the data-transfer cost decreases by at least a factor of the number of mini-batches, which is achieved when a model synchronization is done every epoch.

\subsection{Implementation}\label{sec:sys-impl}

\n{}\footnote{Our code is available at \url{https://github.com/xhuang2016/SDT-GNN}.} is built on top of PyTorch~\cite{NEURIPS2019_9015} and DGL~\cite{wang2019dgl}. For graph partitioning, in addition to \algoname{}, we also implement the existing streaming-based algorithms~\cite{xie2014distributed, gonzalez2012powergraph,petroni2015hdrf,mayer2022out} and integrate them into \n{}. We use the Graph Store Server and Samplers implemented by DGL to store partitioned subgraphs with feature data and sample the subgraphs for the generation of mini-batches, respectively. We implement the distributed training module based on the \texttt{distributed} package provided by PyTorch for model synchronization and data parallel training, where the model averaging is implemented via the \texttt{all\_reduce} primitive.  We integrate several popular GNN models~\cite{kipf2016semi,velivckovic2017graph,hamilton2017inductive,wu2019simplifying,chiang2019cluster,zeng2019graphsaint} into \n{} for distributed GNN training. In addition to the built-in streaming partitioning algorithms and GNN models in \n{}, we implement custom interfaces such that any user-defined streaming partitioning algorithms can be used in \n{} to partition the graph for GNN training purposes, and different new GNN models can also be trained via \n{} under distributed environments.

\section{Evaluation}\label{sec:eval}

In this section, we present extensive experiment results to demonstrate the effectiveness and efficiency of \n{} together with our novel streaming-based partitioning algorithm \algoname{}.

\subsection{Setup}\label{sec:setup}

\noindent \textbf{Hardware.} All experiments are conducted on three Linux servers. The first one has 96~GB RAM and two NVIDIA RTX A6000 48-GB GPUs. The second one has 256~GB RAM and four NVIDIA RTX A6000 48-GB GPUs. The third one has 128~GB RAM and eight NVIDIA V100 16-GB GPUs. We run the experiments on all three servers and observe that their results are identical. 

\begin{table}[t]
\renewcommand{\arraystretch}{1.2}
\caption{Dataset statistics}
\vspace{-1mm}
\label{table:dataset}
\centering
\scriptsize
\begin{adjustbox}{width=\columnwidth,center}
    \begin{tabular}{|c|c|c|c|c|c|c|}
        \hline
        Dataset & \# Nodes & \# Edges & \# Features & \# Classes & Train / Val / Test \\
        \hline
        \hline
        Flickr & 89,250 & 899,756 & 500 & 7 & 0.50 / 0.25 / 0.25 \\
        \hline
        OGB-Arxiv & 169,343 & 2,315,598 & 128 & 40 & 0.54 / 0.29 / 0.17 \\
        \hline
        Yelp & 716,847 & 13,954,819 & 300 & 100 & 0.75 / 0.10 / 0.15 \\
        \hline
        OGB-Products & 2,449,029 & 61,859,140 & 100 & 47 & 0.08 / 0.02 / 0.90 \\
        \hline
        Reddit & 232,965 & 114,615,892 & 602 & 41 & 0.66 / 0.10 / 0.24 \\	
        \hline
        Amazon & 1,569,960 & 264,339,468 & 200 & 107 & 0.85 / 0.05 / 0.10 \\
        \hline
        OGB-Papers & 111,059,956 & 3,231,371,744 & 128 & 172 & 0.01 / 0.001 / 0.002 \\
        \hline
    \end{tabular}
\end{adjustbox}
\end{table}

\vspace{1mm}
\noindent \textbf{Datasets.} We consider seven large real-world graph datasets from DGL~\cite{wang2019dgl} and OGB~\cite{hu2020open}, whose statistics are summarized in Table~\ref{table:dataset}. We convert directed graphs into undirected ones by adding reverse edges, as done in DGL and DistDGL. 

\vspace{1pt}
\noindent \textbf{Baselines.} We consider four state-of-the-art streaming partitioning algorithms such as DBH~\cite{xie2014distributed}, PowerGraph~\cite{gonzalez2012powergraph}, HDRF~\cite{petroni2015hdrf}, and 2PS-L~\cite{mayer2022out}. We use three popular GNNs such as GCN~\cite{kipf2016semi}, GAT~\cite{velivckovic2017graph}, and GraphSAGE~\cite{hamilton2017inductive} for performance evaluation of \n{}. We train each model for 100 epochs and report the test accuracy using the model trained with the best validation accuracy. We run the experiments 10 times and report the average values.

\begin{figure}[t]
    \captionsetup[subfloat]{captionskip=1pt}
    \vspace{0mm}
    \centering
        \includegraphics[width=0.9\linewidth, trim=0cm 0cm 0cm 0cm, clip]{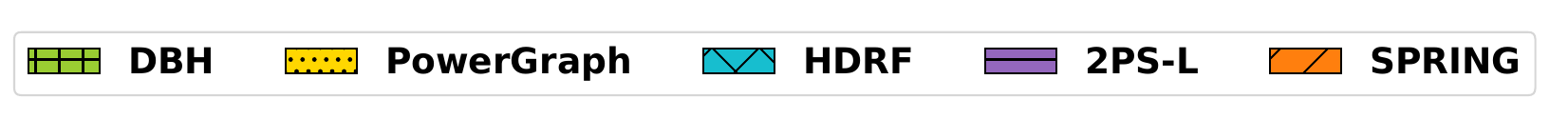}
    \\
    \vspace{-5mm}
    \subfloat[Flickr]{%
        \includegraphics[width=0.32\linewidth, trim=0cm 0cm 0cm 0cm, clip]{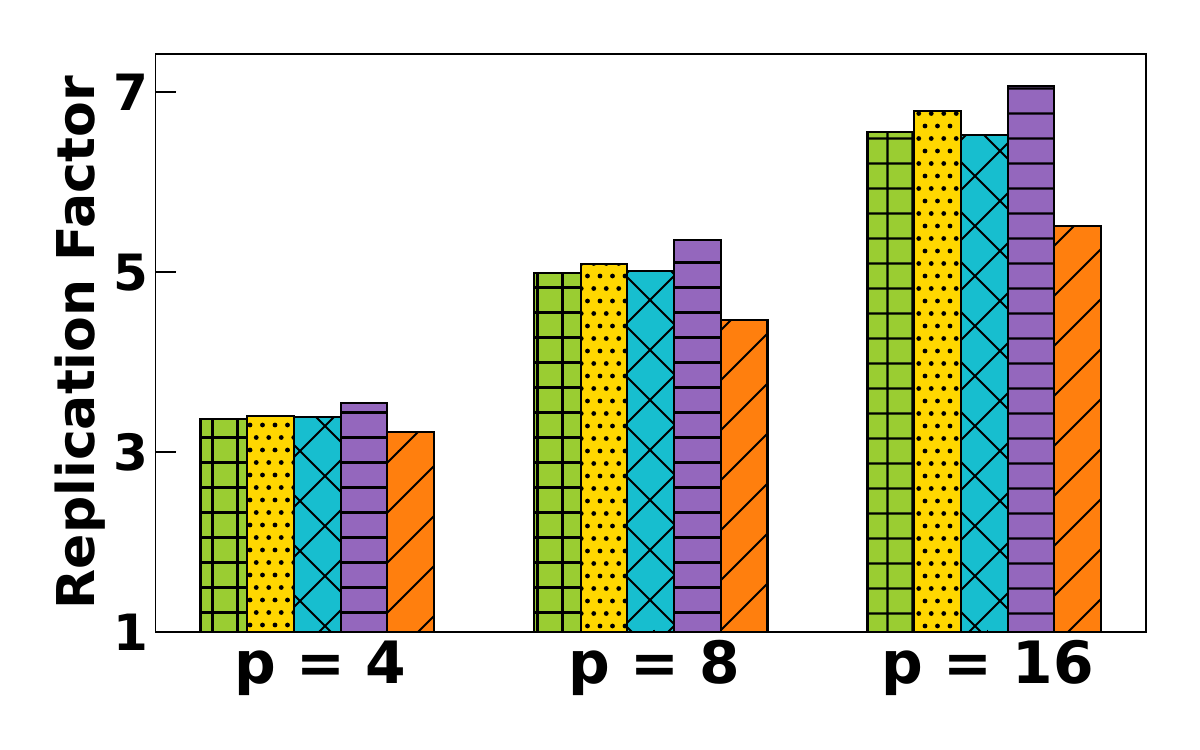}
    }
    \hspace{-2mm}
    \subfloat[OGB-Arxiv]{%
        \includegraphics[width=0.32\linewidth, trim=0cm 0cm 0cm 0cm, clip]{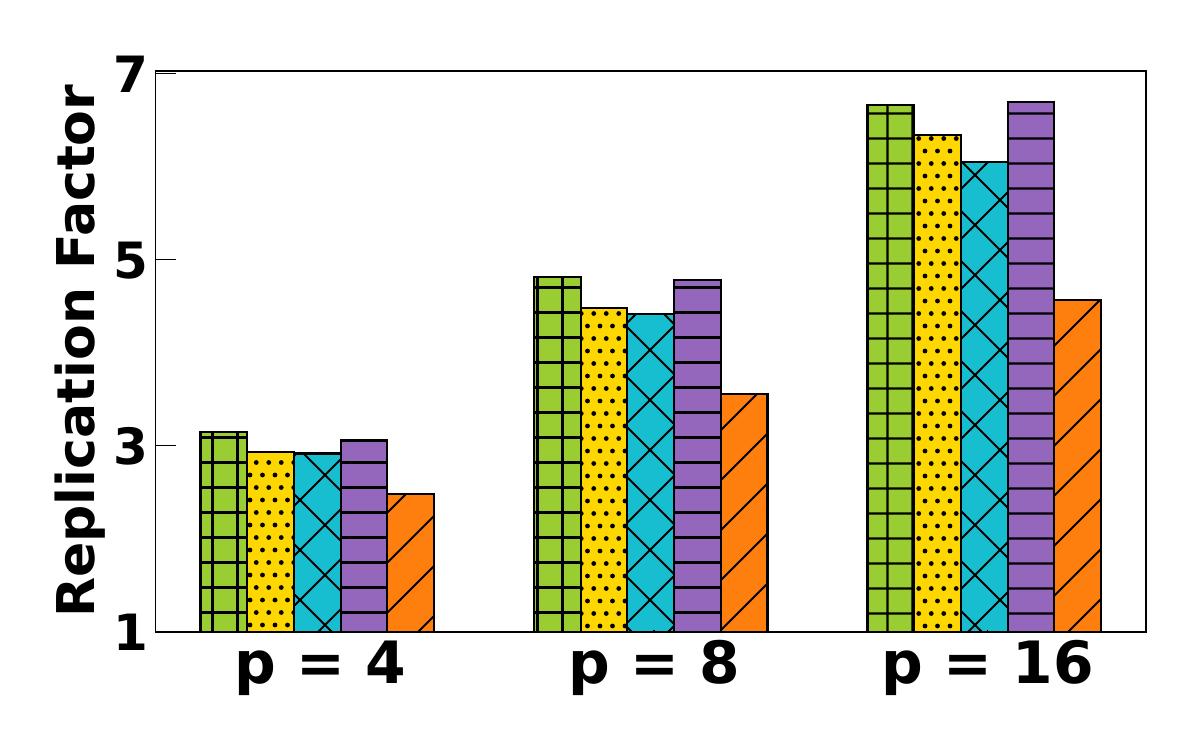}
    }
    \hspace{-2mm}
    \subfloat[Yelp]{%
        \includegraphics[width=0.32\linewidth, trim=0cm 0cm 0cm 0cm, clip]{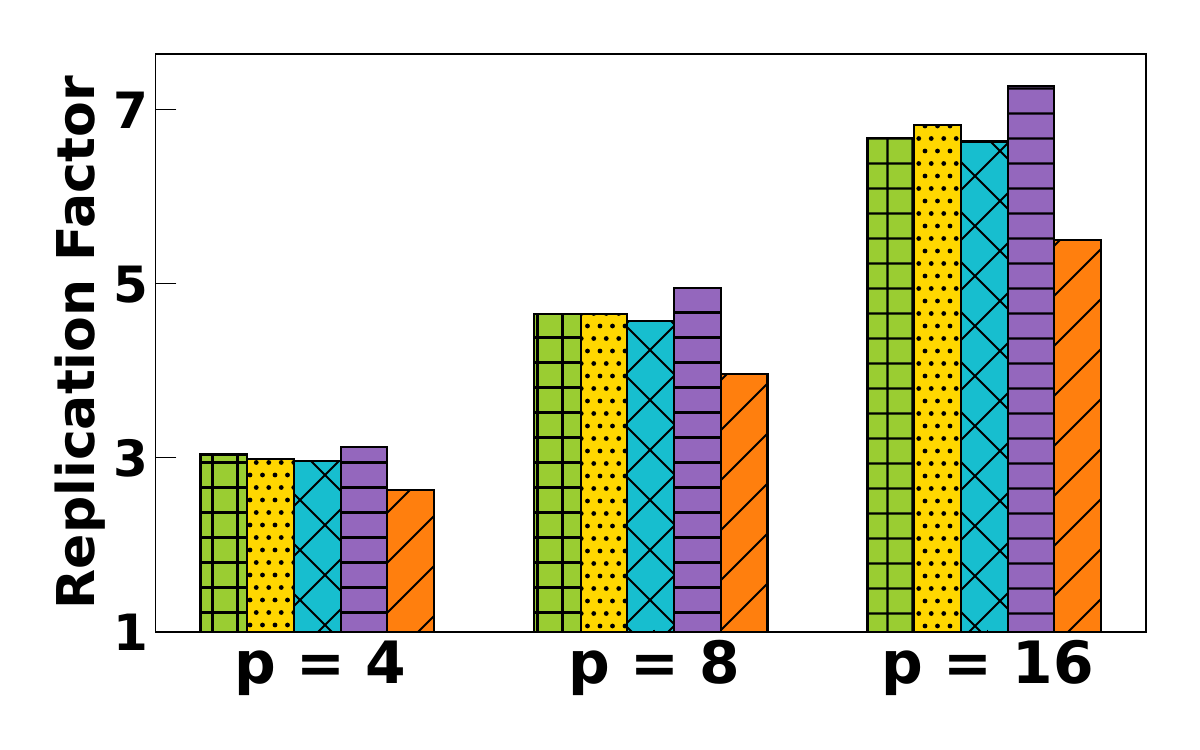}
    }
    \vspace{0mm}
    \subfloat[OGB-Products]{%
        \includegraphics[width=0.32\linewidth, trim=0cm 0cm 0cm 0cm, clip]{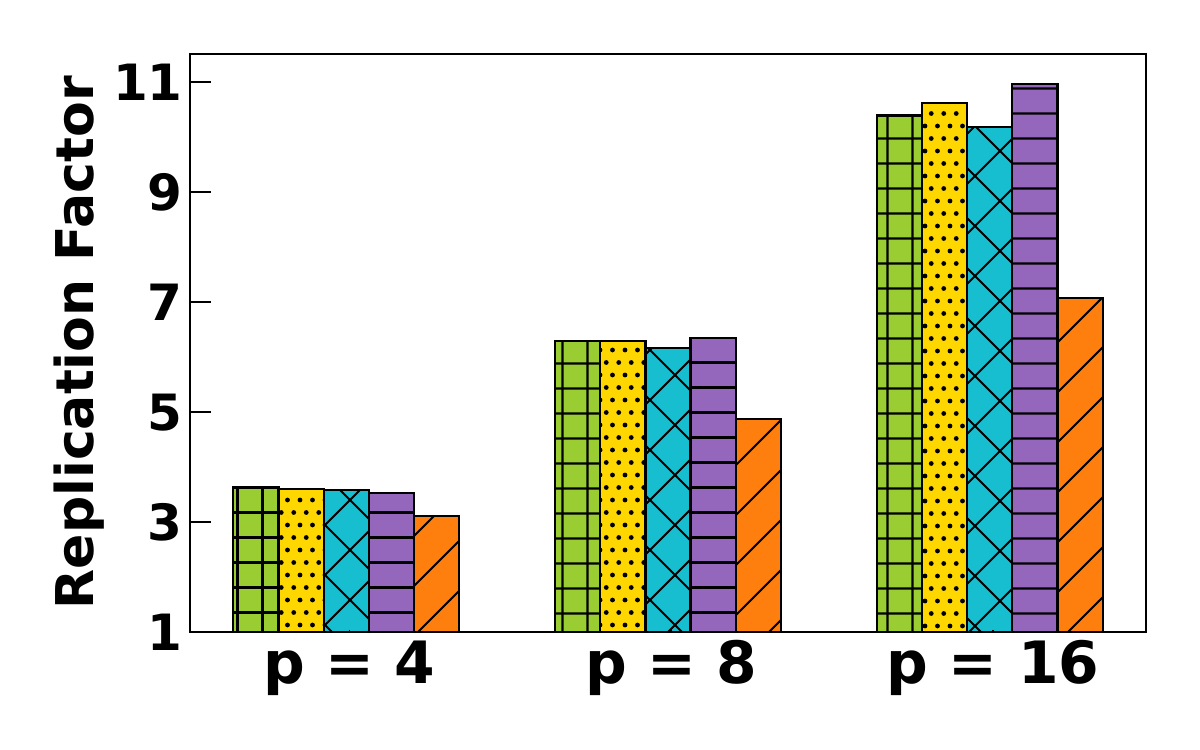}
    }
    \hspace{-2mm}
    \subfloat[Reddit]{%
        \includegraphics[width=0.32\linewidth, trim=0cm 0cm 0cm 0cm, clip]{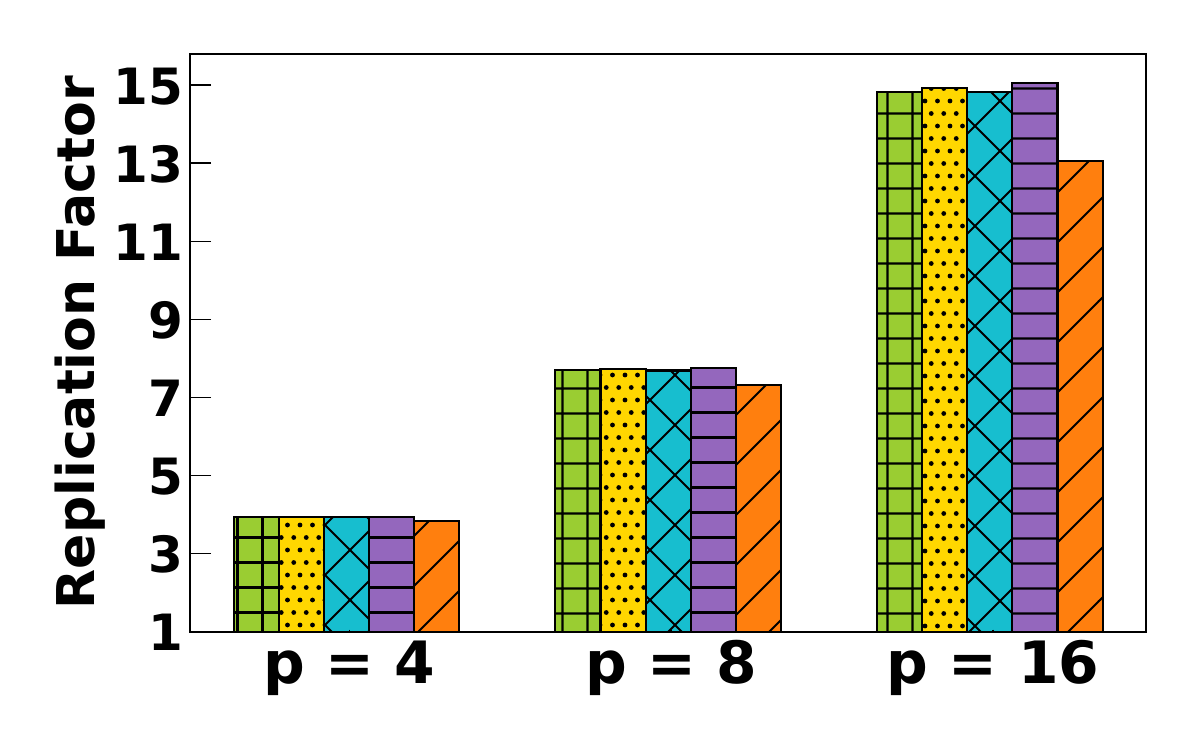}
    }
    \hspace{-2mm}
    \subfloat[Amazon]{%
        \includegraphics[width=0.32\linewidth, trim=0cm 0cm 0cm 0cm, clip]{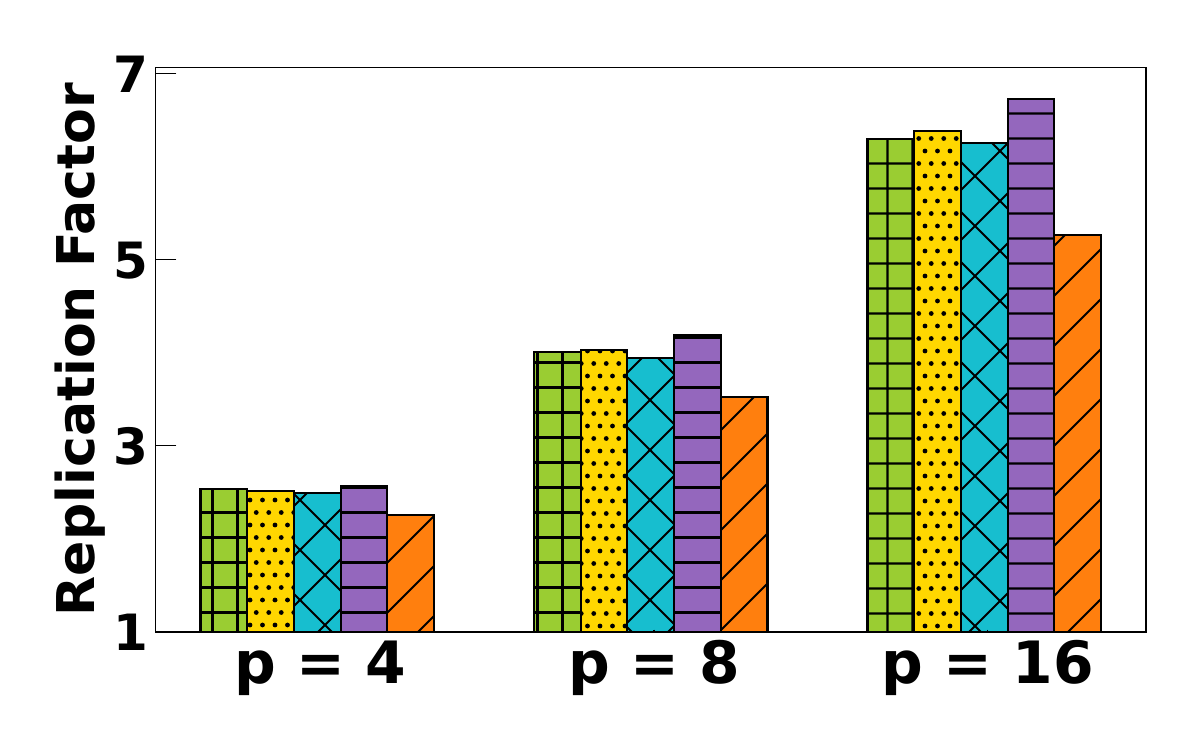}
    }
    \vspace{0mm}
    \caption{Replication factors of partitioning algorithms.}
    \label{fig:RF}
    \vspace{-2mm}
\end{figure}

\begin{figure}[t]
    \centering
    \includegraphics[width=0.95\linewidth, trim=0mm 0mm 0mm 0mm, clip]{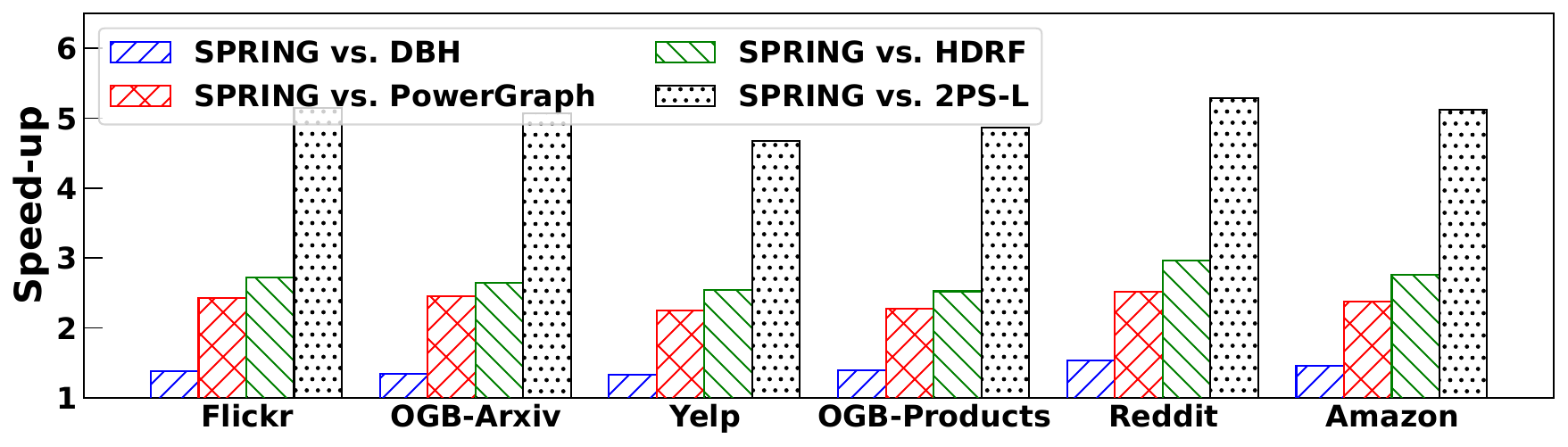}
    \vspace{-3mm}
    \caption{Speed-up by SPRING over the baseline streaming partitioning algorithms when $p=4$.}
    \vspace{0mm}
    \label{fig:Speedup}
\end{figure}

\subsection{Streaming Partitioning Algorithms}\label{sec:eval-alg}

We first demonstrate the effectiveness of \algoname{} compared to the baseline streaming partitioning algorithms in terms of the replication factor. Recall that all the baseline streaming partitioning algorithms were not originally designed for distributed GNN training, but \n{} allows them to be used for that purpose. We partition the graphs into $p \!\in\! \{4,8,16\}$ partitions using \algoname{} and the baselines. The replication factor is defined as $RF_{\mathcal{P}} = ({\sum_{i=1}^p|V_i|})/{|V|}$, where $\mathcal{P} \!=\! \{\mathcal{P}_1, \cdots, \mathcal{P}_p\}$ is a partition result, and $V_i$ is the set of nodes (vertices) in partition $\mathcal{P}_i$. The smaller the replication factor, the smaller the number of node replicas, which leads to the lower memory space needed for GNN-related operations. As shown in Figure~\ref{fig:RF}, we observe that \algoname{} outperforms all the other baselines with an average improvement of about 20\% in the replication factor across all datasets for all different numbers of partitions.

We also demonstrate the efficiency of \algoname{} compared to the baselines in terms of the running time. In Figure~\ref{fig:Speedup}, we present the speed-up ratio of \algoname{} compared to each baseline when $p \!=\! 4$. We observe that \algoname{} can partition each input graph in a few seconds to a few minutes and is faster than all the other baselines for all testing cases. In particular, the speed-up ratio is up to $5\times$, which is achieved when compared to the recent 2PS-L algorithm. This is because the key operations of \algoname{} just rely on the degree and connectivity information, without the need of computing the complicated score functions as done in 2PS-L. All these results indicate that \algoname{} is highly effective and efficient.

\subsection{GNNs Trained Using \n{}}

We next compare the performance of GNNs trained via \n{} with those trained in a centralized manner, i.e., the models are trained based on the original graphs without graph partitioning. Note that the centralized GNN training is much slower than its distributed counterpart. It can also fail to train GNN models on large graphs, e.g., the OGB-Papers dataset, due to their sheer size. Nonetheless, we use the results from the centralized GNN training as reference results for validating the correctness of \n{}, as long as the models can be trained in both ways.

\begin{figure}[t]
\captionsetup[subfloat]{captionskip=1pt}
\centering
\vspace{-0mm}
    \includegraphics[width=0.8\linewidth, trim=0cm 0cm 0cm 0cm, clip]{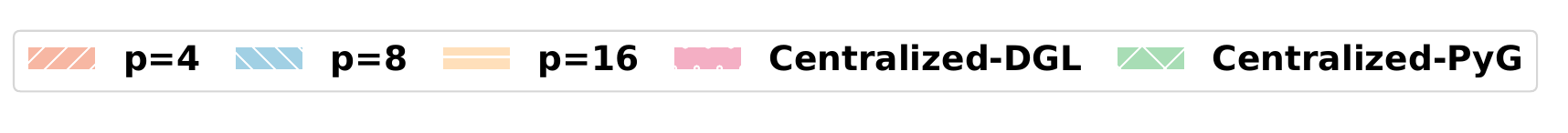}
    \\
    \vspace{-2mm}
    \includegraphics[width=0.98\linewidth, trim=0cm 0cm 0cm 0cm, clip]{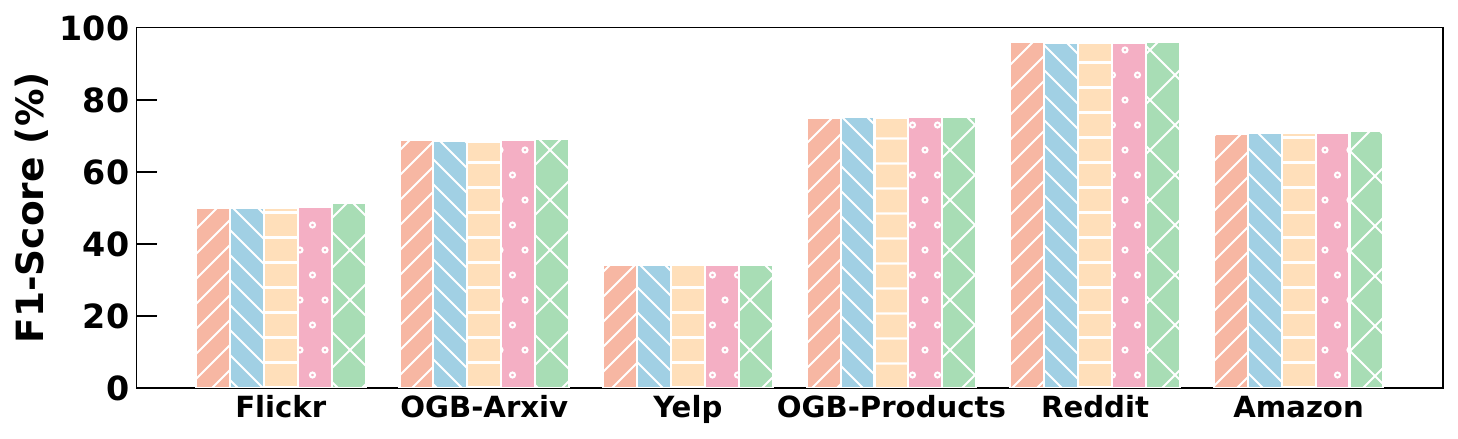}
    \vspace{-3mm}
    \caption{Accuracy of GraphSAGE model trained by \n{} vs. centralized training.}
\label{fig:gnn_acc}
\vspace{-2mm}
\end{figure}

\begin{figure}[t]
    \vspace{0mm}
    \centering
    \includegraphics[width=0.98\linewidth, trim=0mm 0mm 0mm 0mm, clip]{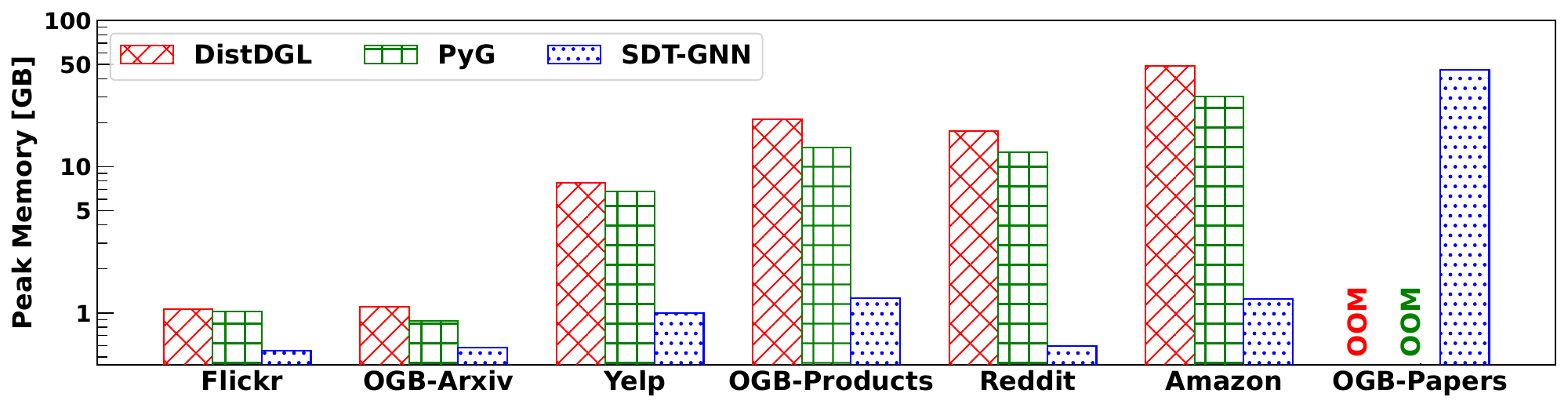}
    \vspace{-3mm}
    \caption{Memory usage of DistDGL, PyG, and \n{}.}
    \vspace{-3mm}
    \label{fig:memory}
\end{figure}

\begin{table}[t!]
\renewcommand{\arraystretch}{1.2}
    \caption{Running time for a training epoch in seconds}
    \vspace{-1mm}
    \label{table:epoch_time}
    \centering
    \footnotesize
    \begin{adjustbox}{width=0.99\columnwidth,center}
        \begin{tabular}{|c|c|c|c|c|c|c|}
        \hline
        & Flickr & OGB-Arxiv & Yelp & OGB-Products & Reddit & Amazon \\
        \hline
        \hline
        DistDGL & 1.855 & 2.372 & 28.549 & \textbf{8.887} & 12.915 & 65.141 \\
        \hline
        PyG & 1.706 & \textbf{1.381} & 35.810 & 13.599 & 21.910 & 75.525 \\
        \hline
        \n{} & \textbf{1.514} & 2.712 & \textbf{27.783} & 9.978 & \textbf{11.994} & \textbf{39.307} \\
        \hline
        \end{tabular}
    \end{adjustbox}
\end{table}

We report the results of GraphSAGE in Figure~\ref{fig:gnn_acc}. We observe that for all testing cases, the GNNs trained by \n{} achieve the same level of accuracy compared to the ones obtained by the centralized training (via DGL and PyG). The results confirm the correctness of \n{}, whose core components include the use of a streaming partitioning algorithm for graph partitioning and the model-averaging strategy for model synchronization among GPUs. We have also instantiated \n{} with other streaming partitioning algorithms and used them to train GNNs. The trained models all obtain similar accuracy to the centralized ones.

\subsection{Comparison with DistDGL and PyG}\label{sec:distdgl}

We turn our attention to the comparison of \n{} against DistDGL and PyG to demonstrate its efficiency, as DistDGL and PyG are two most popular frameworks that are publicly available and well-maintained. We first compare the memory usages of the partitioning modules in \n{}, DistDGL, and PyG. We report the results of the peak memory usage on large graph datasets when partitioning the input graph into four partitions using \algoname{} in \n{} and METIS in DistDGL and PyG. Note that DistDGL and PyG also load the feature data when partitioning the graph, whereas \n{} does not. As shown in Figure~\ref{fig:memory}, we observe that the memory usage of \n{} can be up to $97\%$ and $95\%$ \emph{less} than those of DistDGL and PyG, respectively. In particular, \n{} can handle one of the largest public datasets, OGB-Papers, with about \emph{$85\%$ less} memory footprint than DistDGL and PyG, as they require about 400~GB RAM~\cite{dgl_metis}. This observation shows that \algoname{} can partition the graphs with significantly less memory and thus makes distributed GNN training on large graphs feasible with commodity workstations.

We next evaluate the training speed of \n{} when compared to DistDGL and PyG, although improving the training speed is not the main focus of this paper. We measure the running time for a training epoch of DistDGL, PyG, and \n{} and report the results in Table~\ref{table:epoch_time} for the case of $p\!=\!4$ on the 4-GPU server. Note that we focus on the cases when $p$ is the same as the number of GPUs, since DistDGL and PyG do not support the cases when $p$ is greater than the number of GPUs. We observe that \n{} shows a similar level of training speed as the ones of DistDGL and PyG for most testing cases and is even faster on a few large graphs. That is, \n{} does not sacrifice the per-epoch training speed while highly improving memory efficiency.

\begin{figure}[t]
    \captionsetup[subfloat]{captionskip=1pt}
    \centering
    \vspace{0mm}
    \subfloat[GCN]{%
        \includegraphics[width=0.31\linewidth, trim=0cm 0cm 0cm 0cm, clip]{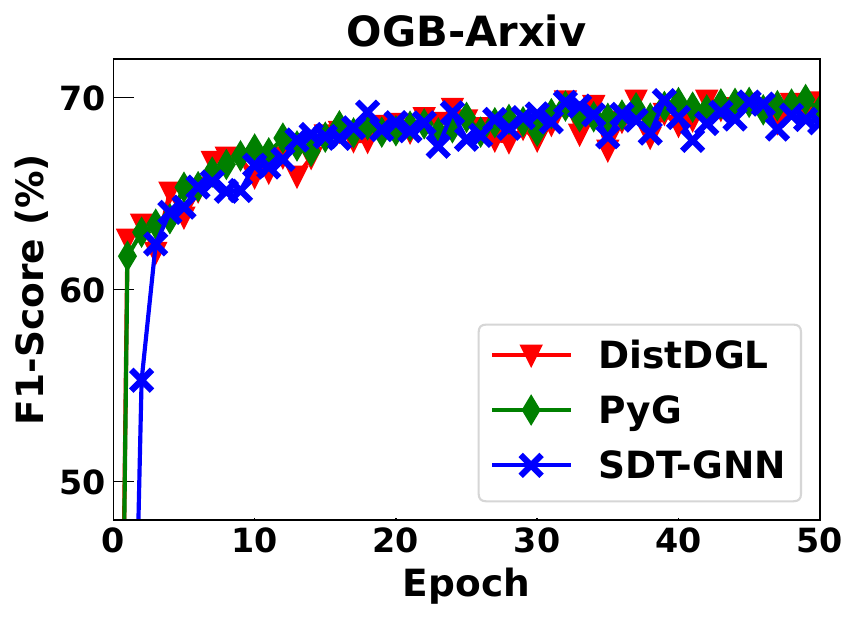}
    }
    \hspace{-1.5mm}
    \subfloat[GAT]{%
        \includegraphics[width=0.31\linewidth, trim=0cm 0cm 0cm 0cm, clip]{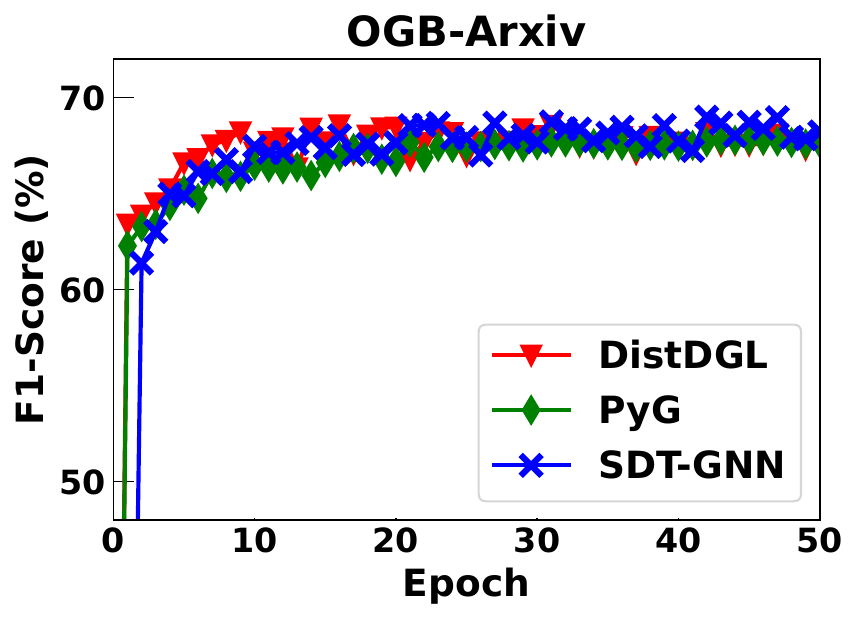}
    }
    \hspace{-1.5mm}
    \subfloat[GraphSAGE]{%
        \includegraphics[width=0.31\linewidth, trim=0cm 0cm 0cm 0cm, clip]{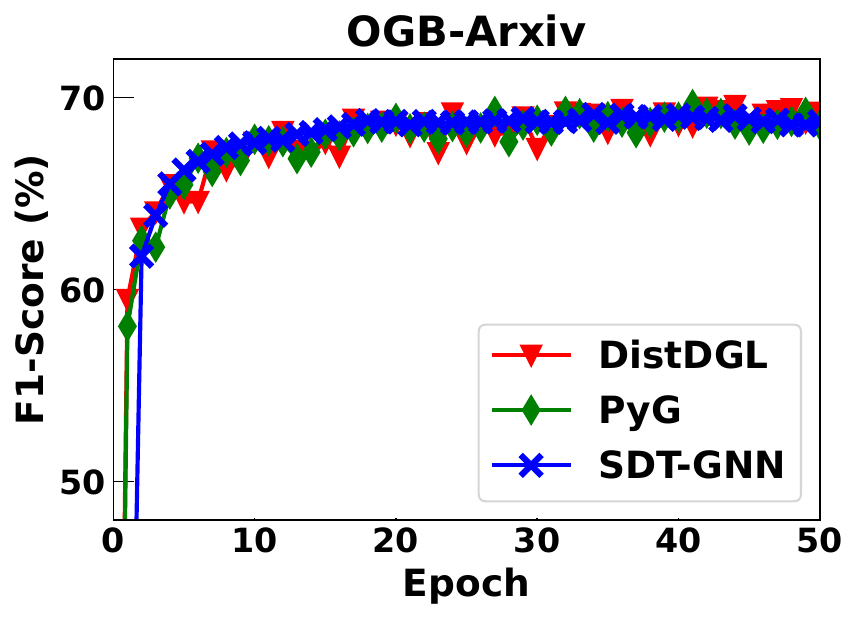}
    }
    \vspace{1mm}
    \subfloat[GCN]{%
        \includegraphics[width=0.31\linewidth, trim=0cm 0cm 0cm 0cm, clip]{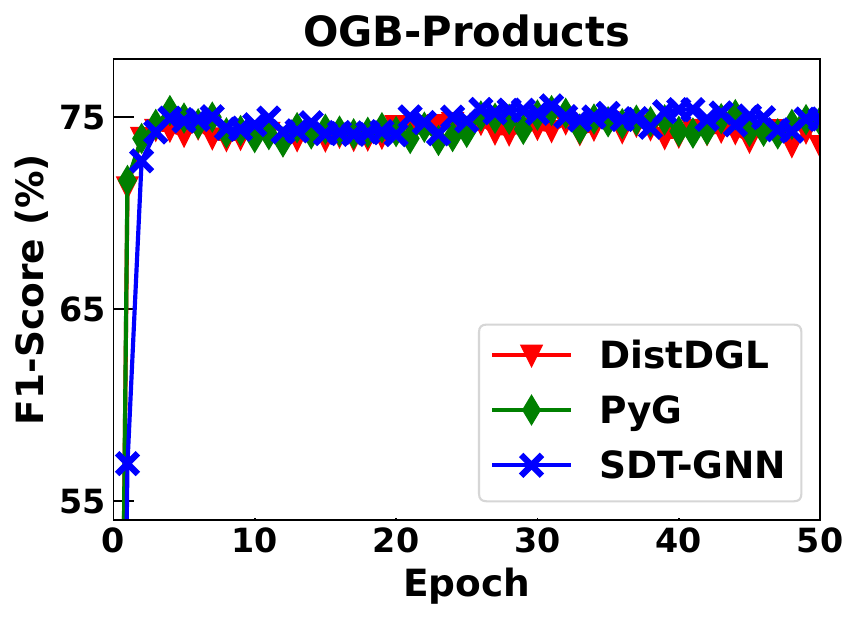}
    }
    \hspace{-1.5mm}
    \subfloat[GAT]{%
        \includegraphics[width=0.31\linewidth, trim=0cm 0cm 0cm 0cm, clip]{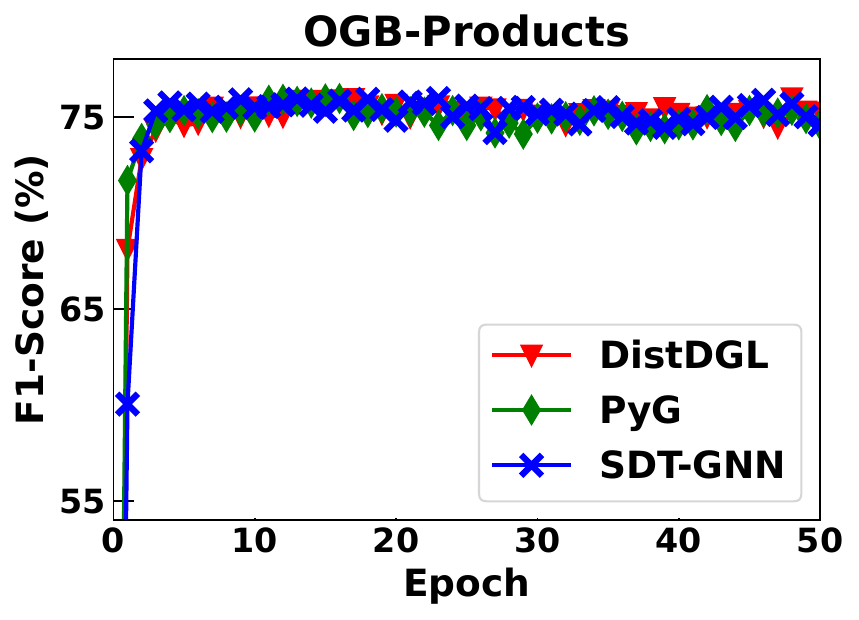}
    }
    \hspace{-1.5mm}
    \subfloat[GraphSAGE]{%
        \includegraphics[width=0.31\linewidth, trim=0cm 0cm 0cm 0cm, clip]{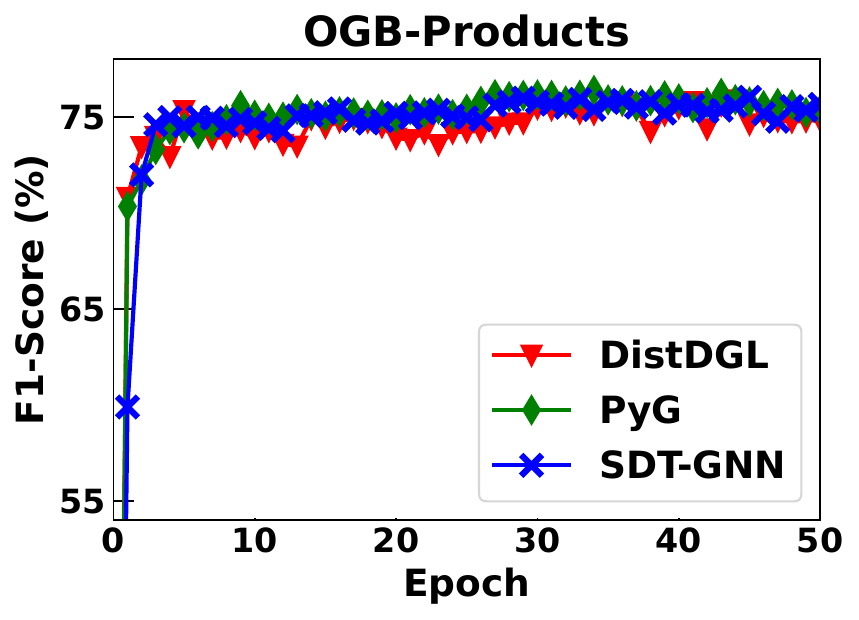}
    }
    \vspace{0mm}
    \caption{F1-scores on OGB-Arxiv and OGB-Products.}
    \label{fig:cvg}
    \vspace{-2mm}
\end{figure}

\begin{figure}[t]
    \captionsetup[subfloat]{captionskip=1pt}
    \centering
    \vspace{0mm}
    \subfloat[2-layer GraphSAGE]{%
        \includegraphics[width=0.42\linewidth, trim=0cm 0cm 0cm 0cm, clip]{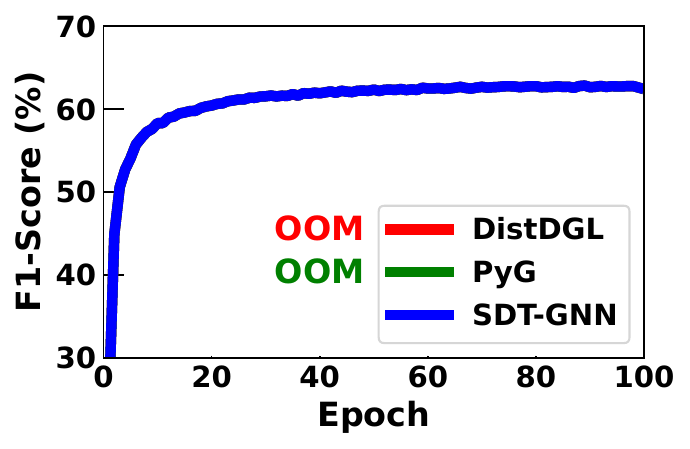}
    }
    \hspace{3mm}
    \subfloat[3-layer GraphSAGE]{%
        \includegraphics[width=0.42\linewidth, trim=0cm 0cm 0cm 0cm, clip]{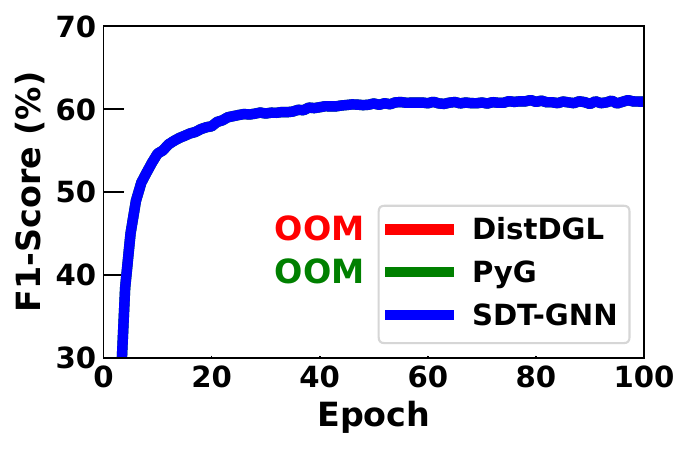}
    }
    \vspace{0mm}
    \caption{F1-scores of GraphSAGE models on OGB-Papers.}
    \label{fig:papers}
\end{figure}

We also compare the accuracy of GNN models trained by DistDGL, PyG, and \n{}. As illustrative results, we present the test accuracy results on the OGB-Arxiv and OGB-Products datasets in Figure~\ref{fig:cvg}. We observe that the test accuracy of all three GNN models trained by \n{} gets improved slowly compared to DistDGL and PyG at first very few epochs. However, it quickly ramps up and eventually converges to almost the same testing accuracy as the ones by DistDGL and PyG within 10 epochs, showing the effectiveness of our distributed training framework. Note that the slow start is due to the fact that the model synchronization in \n{} is done only once per epoch, while DistDGL and PyG synchronize the gradients $n$ times per epoch, where $n$ is the number of mini-batches.

Furthermore, we present the performance of GraphSAGE models trained via \n{} on the OGB-Papers dataset, which \emph{cannot} be handled by DistDGL and PyG on our server with 96~GB RAM, as they require about 400~GB RAM for partitioning~\cite{dgl_metis}. Specifically, we partition the graph into 32 partitions so that each partition can be processed by a GPU for neighborhood sampling and GNN computations. We report the test accuracy results of a 2-layer model and a 3-layer model in Figure~\ref{fig:papers}. The accuracy is comparable to the ones reported in~\cite{zheng2022distributed} and~\cite{kaler2022accelerating}, which leverage the machines with 824~GB RAM and 384~GB RAM, respectively. 

\section{Related Work}
\label{sec:related}

\subsection{Distributed GNN Training}

Most distributed GNN training frameworks and systems are developed based on data parallelism. AGL~\cite{zhang2020agl}, Dorylus~\cite{thorpe2021dorylus}, and DistGNN~\cite{md2021distgnn} are proposed based upon a cluster of CPUs for training, although model training on large graphs via CPUs is deemed inefficient. Several multi-GPU accelerated training systems~\cite{FeyLenssen2019,ma2019neugraph,lin2020pagraph, zhang20212pgraph,jia2020improving,zheng2020distdgl, zheng2022distributed,cai2021dgcl,kaler2022accelerating,kaler2023communication,yang2022wholegraph} are also developed to speed up GNN training. For example, NeuGraph~\cite{ma2019neugraph} and ROC~\cite{jia2020improving} focus on full-batch training, which can hardly be scaled to large graphs. PaGraph~\cite{lin2020pagraph}, 2PGraph~\cite{zhang20212pgraph}, and BGL~\cite{liu2023bgl} aim to accelerate GNN training by reducing the feature data movement overhead via caching strategies. In addition, unlike the others, $P^3$~\cite{gandhi2021p3} leverages a combination of model parallelism and data parallelism to reduce the feature data movement traffic, but it can suffer from a high extra synchronization overhead due to its hybrid parallelism design.

While having their own advantages, all the above distributed training frameworks and systems assume that the memory of the master node is large enough to hold the \emph{entire} graph and use conventional algorithms for graph partitioning. In Table~\ref{table:related}, we summarize the partitioning algorithms used in the existing distributed GNN frameworks and systems. For example, the mainstream distributed GNN training frameworks such as DistDGL~\cite{zheng2020distdgl, zheng2022distributed} and PyG~\cite{FeyLenssen2019} as well as several other systems~\cite{ma2019neugraph,zhang20212pgraph,cai2021dgcl,ramezani2022learn} adopt METIS~\cite{karypis1998multilevelk,karypis1997metis} for graph partitioning. They, however, cannot handle the graphs when their graph data size is too big to fit into the memory. In contrast, \n{} takes a stream of edges as input, without loading the entire graph in memory, and allows existing or new streaming partitioning algorithms to partition graphs for distributed GNN training. In addition, the existing frameworks and systems mainly focus on speeding up distributed GNN training, whereas \n{} is designed from the perspective of improving memory efficiency and scalability. That is, \n{} effectively enables the distributed training of GNNs on large graphs under limited computational resources.

\begin{table}[t!]
\renewcommand{\arraystretch}{1.2}
\caption{Existing distributed GNN training systems}
\vspace{-1mm}
\label{table:related}
\centering
\scriptsize
\begin{adjustbox}{width=\columnwidth,center}
    \begin{tabular}{|c|c|c|}
        \hline
        System & \multicolumn{2}{c|}{Graph Partitioning} \\
        \hline
        & Algorithm & Type \\
        \hline
        \hline
        DistDGL & METIS & in-memory \\
        \hline
        PyG & METIS & in-memory \\
        \hline
        2PGraph & METIS & in-memory \\
        \hline
        DGCL & METIS & in-memory \\
        \hline
        BNS-GCN & METIS & in-memory \\
        \hline
        SALIENT++ & METIS & in-memory \\
        \hline
        AdaQP & METIS & in-memory \\
        \hline
        NeuGraph & METIS, Kernighan-Lin algorithm & in-memory \\
        \hline
        $P^3$ & Random hash algorithm & in-memory \\
        \hline
        WholeGraph & Random hash algorithm & in-memory \\
        \hline 
        AGL & Edge partitioning algorithm & in-memory \\
        \hline
        BGL & Multi-source BFS & in-memory \\
        \hline
        ByteGNN & Multi-source BFS & in-memory \\
        \hline
        DistGNN & Vertex-cut partitioning algorithm & in-memory \\
        \hline
        Dorylus & Edge-cut partitioning algorithm & in-memory \\
        \hline
        PaGraph{\tablefootnote{It adopts a streaming-based partitioning algorithm from \cite{stanton2012streaming} as an `offline' graph partitioning algorithm. Thus, its actual implementation still requires loading the entire graph in memory for partitioning.}} & Greedy algorithm & in-memory \\
        \hline
        ROC & Linear regression model & in-memory \\
        \hline
        \n{} & Edge streaming partitioning algorithms & out-of-core \\
        \hline
    \end{tabular}
\end{adjustbox}
\vspace{1mm}
\end{table}

\subsection{Streaming-based Graph Partitioning}

There have been quite a few streaming-based graph partitioning algorithms~\cite{gonzalez2012powergraph, xie2014distributed, petroni2015hdrf, mayer2022out, jain2013graphbuilder, mayer2018adwise}, although they are not developed for distributed GNN training purposes. For example, PowerGraph~\cite{gonzalez2012powergraph} proposes a greedy algorithm to, for each edge, determine which partition it is assigned to based on where two end nodes of the edge currently belong and the size of their current partition. HDRF~\cite{petroni2015hdrf} improves the greedy algorithm by leveraging the partial node degree information. DBH~\cite{xie2014distributed} proposes a hashing method to cut and replicate high-degree nodes to reduce the replication factor. It is based on the observation that for power-law graphs, replicating high-degree nodes across partitions can significantly reduce the replication factor. Most recently, 2PS-L~\cite{mayer2022out} leverages natural clusters in a graph to improve the partitioning quality. It adopts the streaming clustering algorithm~\cite{hollocou2017streaming} to find clusters and merges the clusters into partitions based on their cluster volume. It further adjusts the partition assignment by maximizing a score function defined based on the cluster volume and node degrees.

We point out that all the streaming partitioning algorithms are originally designed for pure graph mining purposes. They make the neighbors of many nodes fragmented across partitions and thus are not suitable for direct use of distributed GNN training, which requires the full neighbor information. To fill this gap, we develop \n{} to make the existing streaming partitioning algorithms usable for distributed GNN training. We further develop \algoname{} to improve the quality of partitioning, which in turn enables highly effective distributed GNN training with \n{}. 

It is worth noting that there is another class of streaming-based partitioning algorithms~\cite{malewicz2010pregel, stanton2012streaming, stanton2014streaming, tsourakakis2014fennel, tsourakakis2015streaming, zhang2018akin}, which are node-based. They take in a stream of nodes, each of which comes with its neighbor list, as input. Thus, they require the whole graph to be \emph{preprocessed} to obtain the neighbor list of each node beforehand. Due to this reason, we do not consider them in this work for a fair comparison.

\section{Conclusion}

We have developed \n{}, a memory-efficient distributed training framework to train GNNs on large graphs efficiently and effectively. It partitions the input graph with \algoname{} in a streaming manner, automatically distributes the workload to workers, and enables distributed training via model averaging. We have conducted extensive experiments on seven large datasets to demonstrate the effectiveness and efficiency of \n{}. We expect that \n{} can be a cost-efficient distributed framework for training GNNs on large graphs under limited memory and computational resources.

\section*{Acknowledgment}
This work was supported by the National Science Foundation under Grant Nos. 2209921 and 2209922, the International Energy Joint R\&D Program of the Korea Institute of Energy Technology Evaluation and Planning (KETEP), granted financial resource from the Ministry of Trade, Industry \& Energy, Republic of Korea (No. 20228530050030), a grant from SK hynix America, and an equipment donation from NVIDIA Corporation.

\bibliographystyle{IEEEtran}
\bibliography{IEEEabrv,ref}

\end{document}